\documentclass[10pt,twocolumn,letterpaper]{article}

\usepackage{cvpr}              

\usepackage{marvosym}
\usepackage{amssymb}
\usepackage{graphicx}
\usepackage{amsmath}
\usepackage{booktabs}
\usepackage{rotating}
\usepackage{multirow}
\usepackage{enumitem}
\usepackage{algorithmicx,algorithm}
\usepackage{enumitem}
\usepackage{pifont}
\usepackage{makeidx}
\usepackage{epstopdf}
\usepackage[noend]{algpseudocode}
\usepackage[x11names]{xcolor}
\usepackage{colortbl}
\usepackage{color}
\usepackage{makecell}
\usepackage{capt-of}

\definecolor{lightblue}{rgb}{0.761,0.894,0.937} 
\definecolor{white}{rgb}{0.965,0.973,0.988} 
\definecolor{lightpurple}{rgb}{0.914,0.870,0.933} 

\usepackage[skip=0pt]{caption}

\usepackage[pagebackref,breaklinks,colorlinks]{hyperref}
\usepackage[capitalize]{cleveref}
\crefname{section}{Sec.}{Secs.}
\Crefname{section}{Section}{Sections}
\Crefname{table}{Table}{Tables}
\crefname{table}{Tab.}{Tabs.}

\colorlet{dark-blue}{blue!70!black}
\hypersetup{
    colorlinks=true,%
    citecolor=dark-blue,%
    filecolor=dark-blue,%
    linkcolor=red,%
    urlcolor=magenta
}

\usepackage{comment}
\def\cvprPaperID{3166} 
\def\confName{CVPR}
\def\confYear{2026}

\makeatletter
\def\thanks#1{\protected@xdef\@thanks{\@thanks
        \protect\footnotetext{#1}}}
\makeatother

\begin{document}

\title{Open-Vocabulary Domain Generalization in Urban-Scene Segmentation}

\author{Dong Zhao$ ^{1}$, Qi Zang$ ^{2}$ \textsuperscript{\Letter}, Nan Pu$ ^{2}$, Wenjing Li$ ^{2}$, Nicu Sebe$ ^{1}$, Zhun Zhong$ ^{2}$ \textsuperscript{\Letter}\\
$ ^{1}$ Department of Information Engineering and Computer Science, University of Trento, Italy \\
$ ^{2}$ School of Computer Science and Information Engineering, Hefei University of Technology, China \\
}

\makeatletter
\g@addto@macro\@maketitle{
  \begin{figure}[H]
  \vspace{-1.3cm}
  \setlength{\linewidth}{\textwidth}
  \setlength{\hsize}{\textwidth}
  \centering
  \includegraphics[width=1.0\linewidth]{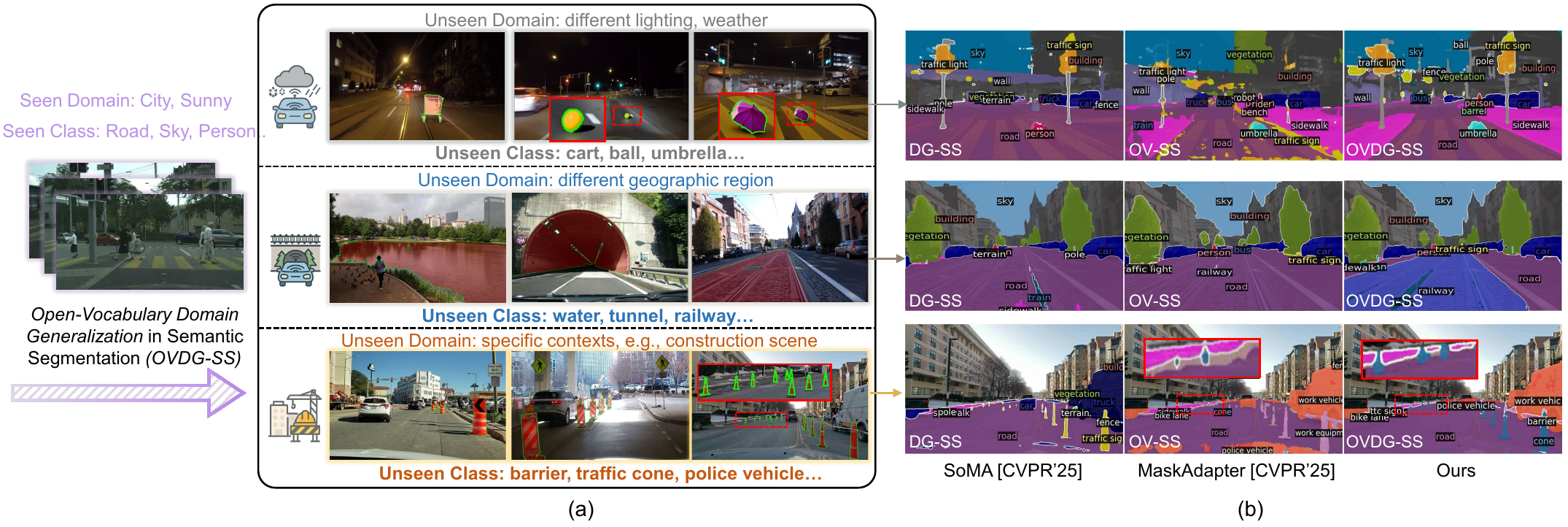}
  \caption{Concept of Open-Vocabulary Domain Generalization in Semantic Segmentation (OVDG-SS). (a) OVDG-SS aims to generalize across both unseen domains (e.g., lighting, weather, geographic region, construction context) and unseen classes (e.g., barrier, cone, railway), moving beyond conventional domain generalization (DG-SS) and open-vocabulary segmentation (OV-SS). (b) Comparison of segmentation results. Traditional DG-SS fails to recognize unseen categories, and OV-SS struggles under domain shifts, while our OVDG-SS effectively handles both unseen domains and classes simultaneously.
}
  \label{Fig1:ovdg_intro}
  \end{figure}
}
\makeatother

\maketitle

\begin{abstract}
Domain Generalization in Semantic Segmentation (DG-SS) aims to enable segmentation models to perform robustly in unseen environments.
However, conventional DG-SS methods are restricted to a fixed set of known categories, limiting their applicability in open-world scenarios.
Recent progress in Vision-Language Models (VLMs) has advanced Open-Vocabulary Semantic Segmentation (OV-SS) by enabling models to recognize a broader range of concepts.
Yet, these models remain sensitive to domain shifts and struggle to maintain robustness when deployed in unseen environments, a challenge that is particularly severe in urban-driving scenarios.
To bridge this gap, we introduce Open-Vocabulary Domain Generalization in Semantic Segmentation (OVDG-SS), a new setting that jointly addresses unseen domains and unseen categories.
We introduce the first benchmark for OVDG-SS in autonomous driving, addressing a previously unexplored problem and covering both synthetic-to-real and real-to-real generalization across diverse unseen domains and unseen categories.
In OVDG-SS, we observe that domain shifts often distort text–image correlations in pre-trained VLMs, which hinders the performance of OV-SS models.
To tackle this challenge, we propose S$^2$-Corr, a state-space-driven text–image correlation refinement mechanism that can mitigate domain-induced distortions and produce a more  consistent text–image correlation under distribution changes.
Extensive experiments on our constructed benchmark demonstrate that the proposed method achieves superior cross-domain performance and efficiency compared to existing OV-SS approaches.
The code is available at \url{https://github.com/DZhaoXd/s2_corr}.
\end{abstract}

\section{Introduction}
\label{sec:intro}
Domain Generalization in Semantic Segmentation (DG-SS) has long been a key challenge in enabling segmentation models to perform robustly in unseen environments~\cite{zang_style_2025, zang_tnnls_2025}.
The recent emergence of large Vision Foundation Models (VFMs)~\cite{dinov2, eva02} has notably improved the cross-domain robustness of segmentation models~\cite{Wei_2024_CVPR, Zhao_2025_CVPR}.
Despite this progress, these models remain limited to seen semantics~\cite{zhao2023semantic, zhao_pami_2025}, as they can only recognize categories present in their training data~\cite{NEURIPS2024_b78c6928, zheng_TKM}.
This limitation becomes critical for safety in autonomous driving scenarios, where models trained on urban sunny scenes often fail to detect unseen objects such as barriers or traffic cones appearing at night, in tunnels, or under adverse weather conditions, as illustrated in Fig.~\ref{Fig1:ovdg_intro}(a).
Such failures restrict the scalability of DG-SS toward truly open-world perception.

Open-Vocabulary Semantic Segmentation (OV-SS)~\cite{li2022languagedriven, xu2022simplebaseline} leverages Vision-Language Models (VLMs)~\cite{clip, fang2024eva} to recognize diverse visual concepts beyond closed-set categories. Most existing OV-SS models~\cite{li2024densevlm, wysoczanska2023clipdino, zhang2024corrclip}, trained on COCO-Stuff~\cite{COCO_dataset}, achieve strong results on generic scenes but suffer sharp performance drops when transferred to domain-specific scenarios such as remote sensing~\cite{li2025segearthov} or autonomous driving~\cite{ghosh2025roadwork}, even when many classes overlap with the target domain (Table~\ref{Tab1_intro_ovdg}).  
To mitigate this degradation, we further train these OV-SS models on several driving-related datasets under a limited-vocabulary setting. As shown in Table~\ref{Tab1_intro_ovdg}, models trained on domain-related data (colored rows) exhibit progressively improved performance. When the training domain shifts from synthetic (SYNTHIA, GTA) to real (Cityscapes) and becomes closer to the target distribution, performance consistently improves. \textit{This confirms that OV-SS models are highly sensitive to domain shift and struggle to generalize when the training and target domains are mismatched.}

To this end, we introduce Open-Vocabulary Domain Generalization in Semantic Segmentation (OVDG-SS), a new setting that jointly addresses unseen domains and unseen categories.
As illustrated in Fig.~\ref{Fig1:ovdg_intro}, OVDG-SS aims to adapt models to new environments while recognizing novel concepts beyond the training distribution.
This setting extends OV-SS to real-world applications such as autonomous driving, where robustness to domain shifts and openness to new semantics are both critical for safety.

\begin{table}[t]
  \centering
   \renewcommand{\arraystretch}{1.05}
  \resizebox{0.485\textwidth}{!}{
\begin{tabular}{ccc|c|cc|cc}
\toprule
\multirow{2}[2]{*}{OV-SS Method} & \multirow{2}[2]{*}{Training Data} & Seen  & \multirow{2}[2]{*}{Num.} & \multicolumn{2}{c|}{Shared classes} & \multicolumn{2}{c}{Performance} \\
      &       & class &       & Dv-19 & Dv-58 & Dv-19 & Dv-58 \\
\midrule
CAT-Seg~\cite{cho2024cat} & \multicolumn{1}{l}{ COCO-Stuff~\cite{COCO_dataset}} & 171   & 118K  & 17    & 34    & 31.6  & 32.4  \\
\rowcolor[rgb]{ .886,  .937,  .855} CAT-Seg & SYNTHIA~\cite{ros2016synthia} & 7     & 9.4K  & 7     & 7     & 43.6  & 45.3  \\
\rowcolor[rgb]{ .776,  .878,  .706} CAT-Seg & GTA~\cite{richter2016GTA5} & 7     & 25K   & 7     & 7     & 47.5  & 48.2  \\
\rowcolor[rgb]{ .663,  .816,  .557} CAT-Seg & Cityscapes~\cite{cordts2016cityscapes} & 7     & 3K    & 7     & 7     & \textbf{49.3}  & \textbf{50.0}  \\
\midrule
MaskAdapter~\cite{Li_2025_CVPR} &  COCO-Stuff & 171   & 118K  & 17    & 34    & 30.1  & 29.6  \\
\rowcolor[rgb]{ .867,  .922,  .969} MaskAdapter & SYNTHIA & 7     & 9.4K  & 7     & 7     & 42.4  & 43.1  \\
\rowcolor[rgb]{ .741,  .843,  .933} MaskAdapter & GTA  & 7     & 25K   & 7     & 7     & 46.5  & 45.6  \\
\rowcolor[rgb]{ .608,  .761,  .902} MaskAdapter & Cityscapes & 7     & 3K    & 7     & 7     & \textbf{50.7} & \textbf{49.3}  \\
\bottomrule
\end{tabular}%
}
 	\setlength{\abovecaptionskip}{0.05 cm}
  \caption{Effect of training domains on training-based OV-SS performance.
Colored rows indicate models trained on driving datasets. 
Dv-19 and Dv-58 refer to cross-domain driving datasets with 19 and 58 categories.}
  \label{Tab1_intro_ovdg}%
  \vspace{-0.3cm}
\end{table}%

To facilitate research in this new setting, we construct the first comprehensive OVDG-SS benchmark tailored for urban-driving segmentation. 
It includes both synthetic-to-real and real-to-real generalization settings and spans three unseen domain types: 
\includegraphics[height=1.4em]{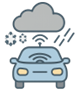} diverse weather and lighting, 
\includegraphics[height=1.4em]{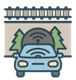} geographically distinct regions, and 
\includegraphics[height=1.4em]{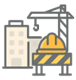} construction-heavy environments. 
Compared to standard DG-SS benchmarks, our benchmark introduces over 30 additional driving-related categories, greatly enriching the semantic space for OVDG evaluation.
Building on this benchmark, we evaluate several representative training-based OV-SS models. 
Although strong under conventional OV-SS settings~\cite{liang2023open}, these models exhibit degradation when transferred to unseen domains, exposing their limited robustness.

\begin{figure}[tbp]
    \begin{center}
    \centering \includegraphics[width=0.4050\textwidth]{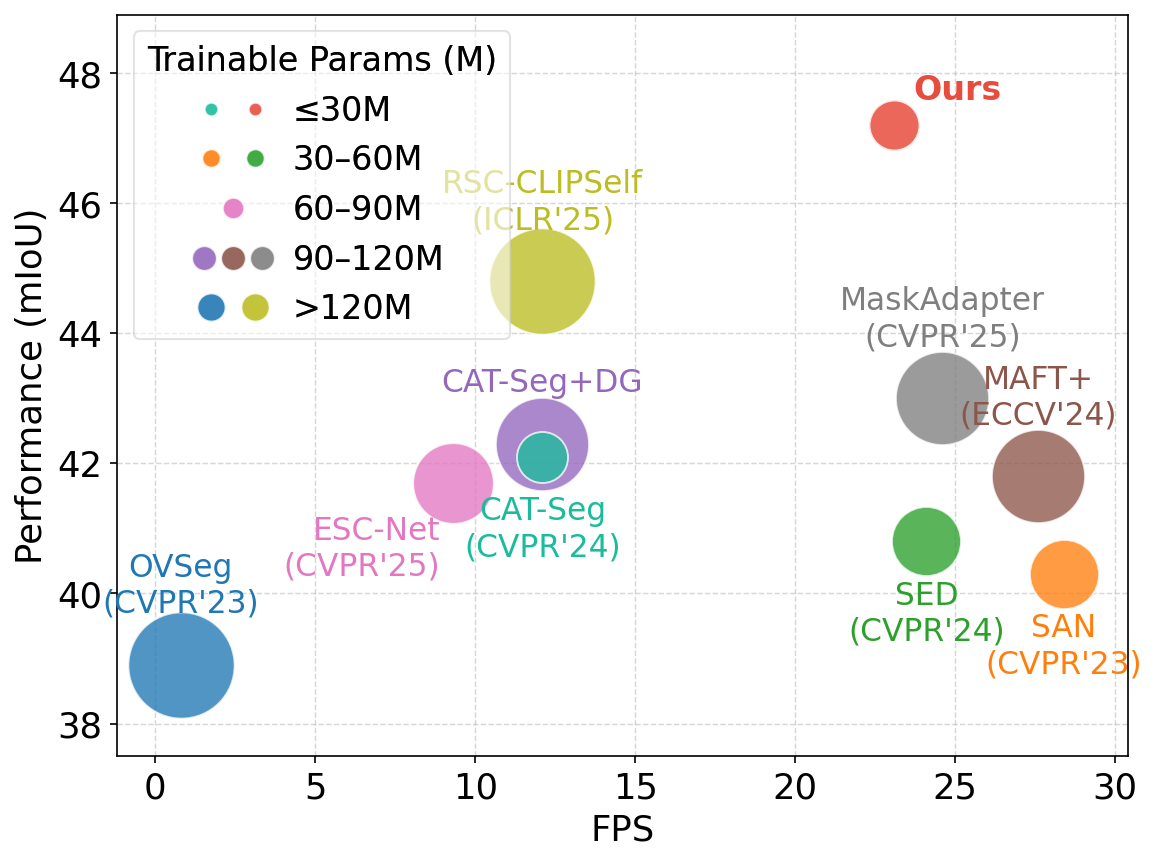}
    \end{center}
    \setlength{\abovecaptionskip}{-0.3 cm}
    \caption{Efficiency and performance comparison on OVDG-SS tasks using EVA02 ViT-B/16 as backbone. 
    FPS is tested on images with a short edge of 480 and a long edge of 960. 
    Our method achieves the best trade-off among generalization ability, speed, and parameter efficiency.} 
    \label{Fig2_Fps}
\vspace{-0.3cm}
\end{figure}

We then analyze why existing OV-SS approaches fail under domain shifts and \textit{find that domain shifts make the initial text–image correlations from VLMs noisy and misaligned}, which in turn severely limits generalization in OVDG-SS.
To address this issue, we propose S$^2$-Corr, a novel \textbf{S}tate-\textbf{S}pace-driven module that dynamically refines noisy text–image \textbf{Corr}elations for robust OVDG-SS.
Built upon a selective state-space~\cite{mamba2} aggregation baseline,  
S$^2$-Corr introduces three key innovations: 
(i) image- and text-conditioned modulation that injects domain-relevant cues, 
(ii) a distance-aware decay mechanism that suppresses long-range noise during sequential aggregation, and 
(iii) a chunk-wise snake scanning strategy that aligns state propagation with the spatial structure. 
Together, these designs reconstruct cleaner and more coherent text-image correlations under distribution shifts, enabling robust open-vocabulary generalization across unseen domains.

As shown in Fig.~\ref{Fig2_Fps}, our method attains higher mIoU and faster inference speed with fewer trainable parameters compared to existing OV-SS approaches, demonstrating both effectiveness and efficiency.
Our main contributions are summarized as follows:
\begin{itemize}
    \vspace{-.05in}
    \item We analyze and empirically verify the limitations of existing DG-SS and OV-SS paradigms, 
    revealing that neither can effectively generalize to both unseen domains and categories simultaneously.
    Based on these findings, we introduce Open-Vocabulary Domain Generalization in Semantic Segmentation (OVDG-SS), a new setting that jointly addresses these challenges.
    \vspace{-.05in}
    \item We construct the first comprehensive benchmark for OVDG-SS in driving scenarios, covering both synthetic-to-real and real-to-real OVDG settings with diverse unseen domains and unseen categories. 
    \vspace{-.05in}
    \item We propose S$^2$-Corr, a novel and efficient state-space-driven correlation refinement module that stabilizes noisy text–image correlations under domain shifts.
    \vspace{-.05in}
   \item S$^2$-Corr achieves the strongest overall performance across unseen domains and categories, providing a solid new baseline for future OVDG-SS research.
\end{itemize}

\begin{figure*}[t]
    \setlength{\abovecaptionskip}{-0.4cm}
    \begin{center}
    \centering 
    \includegraphics[width=0.972\textwidth, height=0.4\textwidth]{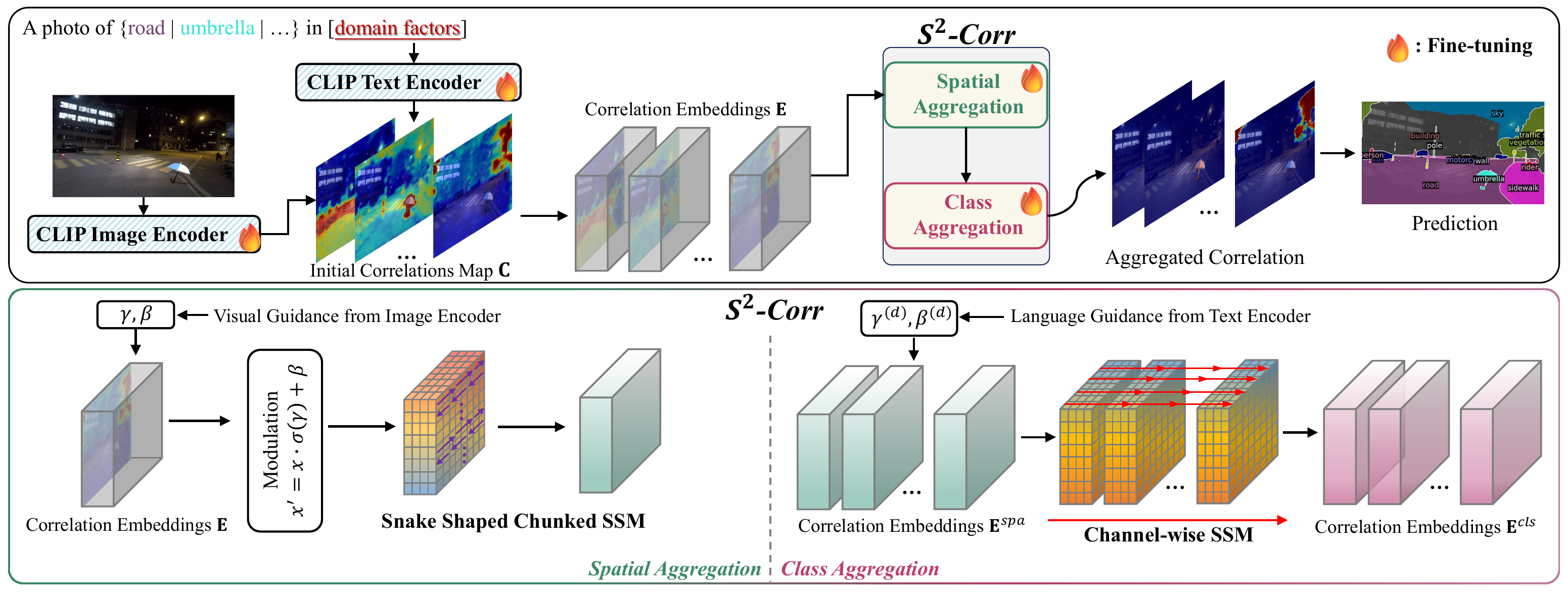} 
    \end{center}
    \caption{Overview of the proposed S$^2$-Corr.
    The upper part shows the CLIP-based encoding and correlation aggregation pipeline. The lower part illustrates our S$^2$-Corr, which refines text–image correlations using a specially designed chunked State-Space Models (SSM) aggregation scheme.} 
    \label{Fig3_pipeline}
    \vspace{-0.3cm}
\end{figure*}

\section{Related Work}
\noindent\textbf{Domain Generalized Semantic Segmentation} (DGSS) 
methods mainly fall into two categories. 
Data augmentation approaches increase domain diversity through image- ~\cite{zhong2022adversarial,  jia2024dginstyle,  zang_tnnls_2025}or feature-level style perturbations~\cite{zang_style_2025, chattopadhyay2023pasta}, or by incorporating additional diverse data~\cite{Zhao_2025_ICCV, Zang_changediff_2025}. 
The rise of VFMs has driven the development of Parameter-Efficient Fine-Tuning (PEFT) strategies. 
Some methods add lightweight adapters~\cite{wei2024stronger, bi2024learning, yi2024learning}, while others update only a small set of important parameters~\cite{Zhao_2025_CVPR, yun2025soma}, enabling efficient adaptation.
TQDM~\cite{pak2024textual} also leverages VLMs, but its design remains confined to closed-set DG-SS.
Overall, these DG approaches are tailored for VFMs or closed-set settings and do not extend naturally to OVDG-SS.

\noindent\textbf{Open-Vocabulary Semantic Segmentation} (OV-SS)
mainly evolved into two main paradigms.
Training-free methods adapt CLIP to segmentation without pixel-level supervision, 
typically enhancing its dense predictions through refined spatial–text alignment~\cite{luo2023segclip, ranasinghe2023perceptual}, 
self-correction of noisy attention maps~\cite{zhou2022extract, lan2024clearclip, bousselham2024grounding, wang2024sclip}, 
self-distillation~\cite{wu2023clipself, ESCNet_2025_CVPR}, 
or VFM-based distillation that injects stronger dense priors~\cite{wang2025declip, jose2025dinov2}.
Training-based methods enhance OV-SS performance by fine-tuning on generic segmentation datasets such as COCO-Stuff. 
These approaches generate category-agnostic masks and assign text-based labels~\cite{kang2024defense, barsellotti2024training, sun2024clip, shao2024explore}, 
improve mask proposal quality~\cite{xu2023side, xu2023san, liu2024open}, 
or refine cross-modal attention maps more effectively~\cite{cho2024cat, wang2025declip}. 
However, we observe that both paradigms generalize poorly on driving-related OVDG-SS tasks, which motivates us to develop a more robust framework tailored for OVDG.

\begin{figure*}[t]
    \setlength{\abovecaptionskip}{-0.4cm}
    \begin{center}
    \centering 
    \includegraphics[width=0.972\textwidth]{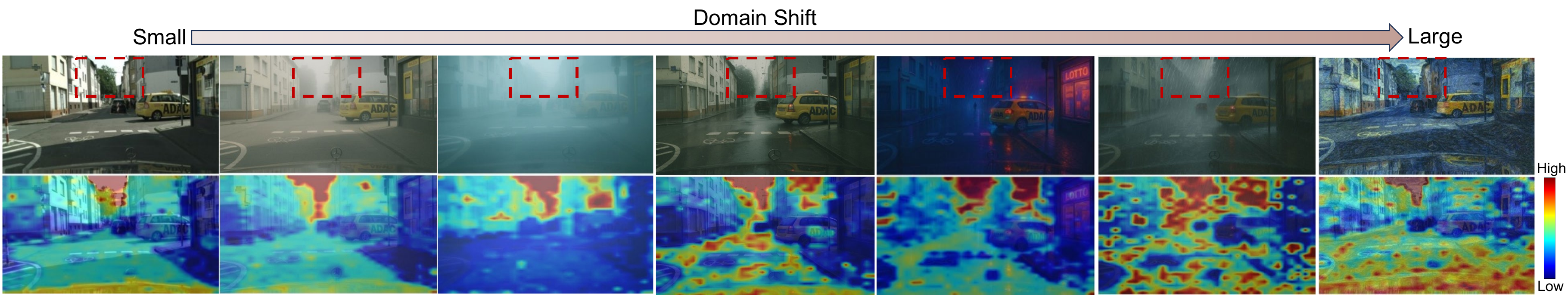} 
    \end{center}
    \caption{Effect of domain shift on text–image correlations of the class ``\textit{sky}'' from EVA02 model~\cite{eva02}. Color map ranges from \textcolor{blue}{\textbf{blue (low correlation)}} to \textcolor{red}{\textbf{red (high correlation)}}. As the domain shift increases from left to right, the initial correlation maps become progressively noisier, with incorrect activations spreading across irrelevant regions.} 
    \label{Fig7_corr_domain_shift}
    \vspace{-0.3cm}
\end{figure*}

\section{Methodology}
\label{sec:method}
\noindent \textbf{Problem Definition.}
We study the Open-Vocabulary Domain Generalized Semantic Segmentation (OVDG-SS) task.  
Given a source domain $\mathcal{D}_{s}=\{(x_i, y_i)\}$ with pixel labels from a base vocabulary of $N_d$ classes,  
the model is deployed on multiple unseen target domains $\{\mathcal{D}_{t}^{(k)}\}_{k=1}^{K}$ that contain a much larger open vocabulary with $M \gg N_d$ classes.  
The objective is to learn a segmentation model that generalizes to unseen domains and recognizes both source classes and novel vocabulary categories.

\noindent\textbf{Overview.} We begin by revisiting existing OV-SS approaches and analyzing why they fail under domain shifts (Sec.~\ref{sec:3.1}).  
Next, we establish a strong baseline by applying a state-space model to aggregate text–image correlations (Sec.~\ref{sec:3.2}).  
Finally, we introduce S$^2$-Corr module and explain how it improves upon this baseline (Sec.~\ref{sec:3.3}). The framework of our method is shown in Fig.~\ref{Fig3_pipeline}.

\subsection{Revisiting and Analyzing OV-SS} \label{sec:3.1}

In this work, we use CAT-Seg~\cite{cho2024cat} as our baseline, owing to its simple correlation-based design and strong robustness in OV-SS. Specifically, given an image–text pair, CAT-Seg extracts visual features 
$\mathbf{F}_v \in \mathbb{R}^{HW \times d}$ 
and textual class embeddings 
$\mathbf{F}_t \in \mathbb{R}^{N_C \times d}$ 
from a vision–language backbone (e.g., CLIP), where $H$ and $W$ are spatial dimensions, 
$d$ is the feature dimension, and $N_C$ is the number of classes. 
The initial correlation map is
\begin{equation}
\mathbf{C} = \text{Norm}(\mathbf{F}_v \mathbf{F}_t^\top) \in \mathbb{R}^{HW \times N_C}.
\label{initial_map}
\end{equation}
To reduce noise and misaligned activations~\cite{wang2024sclip}, CAT-Seg performs a two-stage refinement.

\noindent \textbf{Spatial Aggregation.} To lift the correlation map into a $d_f$-dimensional embedding space, 
CAT-Seg applies a learnable projection 
$\mathbf{P}=[\mathbf{p}_1,\dots,\mathbf{p}_{N_C}] \in \mathbb{R}^{d_f \times N_C}$.
The lifted embeddings are computed by a class-wise broadcasted multiplication:
$
\mathbf{E}_{i,j,:} = C_{ij}\,\mathbf{p}_j,
\mathbf{E} \in \mathbb{R}^{HW \times N_C \times d_f},
$
which can be compactly written as
$
\mathbf{E} = \mathbf{C} \odot \mathbf{P}^\top.
$
Spatial refinement is then applied independently for each class using a shared cross-attention:
\begin{equation}
\mathbf{E}^{\text{spa}} 
= \mathrm{CrossAttn}_{\theta}\!\left(
\mathbf{E},\;
\mathrm{Neigh}(\mathbf{E})
\right),
\label{Eq_spatical_agg}
\end{equation}
where $\theta$ are class-shared parameters and 
$\mathrm{Neigh}(\mathbf{E})$ gathers local spatial neighborhoods 
(e.g., shifted windowed regions as in~\cite{liu2021swin}).
This produces the spatially aggregated embeddings 
$\mathbf{E}^{\text{spa}} \in \mathbb{R}^{HW \times N_C \times d_f}$.

\noindent \textbf{Class-wise Aggregation.}
To model inter-class relationships at each spatial location, 
CAT-Seg applies another cross-attention across the class dimension:
\begin{equation}
\mathbf{E}^{\text{cls}}
=
\mathrm{CrossAttn}^{\text{cls}}_{\phi}
\!\left(
\mathbf{E}^{\text{spa}},
\mathrm{Classes}(\mathbf{E}^{\text{spa}})
\right),
\end{equation}
where $\phi$ are shared parameters and 
$\mathrm{Classes}(\mathbf{E}^{\text{spa}})$ gathers class embeddings at the same spatial position.
The resulting $\mathbf{E}^{\text{cls}} \in \mathbb{R}^{HW \times N_C \times d_f}$  
serves as the final refined correlation map, which is decoded into pixel-wise predictions.

\noindent \textbf{Why do OV-SS method perform poorly in OVDG-SS?}  
Although CAT-Seg performs well in standard OV-SS settings, its performance drops sharply under OVDG-SS. We attribute this to two main factors.
(1) \textit{Heavy correlation map noise induced by domain shift.}
As shown in Fig.~\ref{Fig7_corr_domain_shift}, large domain shifts corrupt the initial text–image correlation map \(\mathbf{C}\), 
resulting in biased and spatially inconsistent activations that hinder downstream correlation refinement.
(2) \textit{Noise propagation via cross-attention.} 
As in Eq.~\ref{Eq_spatical_agg}, CAT-Seg refines class-wise correlation embeddings through attention-based aggregation.
Under domain shift, the correlation map $\mathbf{C}$ contains many corrupted activations, which enter the cross-attention as noisy keys and values and distort the attention weights.
These errors then propagate to neighboring positions within each class, and subsequent class aggregation further amplifies such domain-specific noise.

\subsection{Refining Correlation with State-Space Models} \label{sec:3.2}
To construct a more robust baseline for correlation aggregation, 
we replace the attention-based aggregation in CAT-Seg with a selective state-space model (SSM)~\cite{mamba, mamba2}, 
which processes correlations in a sequential manner. 

For spatial aggregation, given the correlation embeddings $\mathbf{E} \in \mathbb{R}^{H W \times N_C \times d_f}$, 
we flatten the spatial grid into a 1D sequence under a fixed scan order $\pi$ (e.g., row-major):
\begin{equation*}
\mathbf{x}_t = \mathbf{E}_{\pi(t), :} \in \mathbb{R}^{d_f}, 
\qquad t = 1, \dots, T, \; T = H W,    
\end{equation*}
where $\mathbf{x}_t$ represents the correlation embedding at the $t$-th spatial position.
Instead of explicit token-to-token interaction through cross-attention, 
we refine the sequence using selective recurrent state updates governed by a continuous-time state-space model,
\begin{equation}
\mathbf{h}_t = \mathbf{A}_t\,\mathbf{h}_{t-1} + \mathbf{B}_t\,\mathbf{x}_t, 
\qquad
\mathbf{y}_t = \mathbf{W}_t\,\mathbf{h}_t + \mathbf{U}_t\,\mathbf{x}_t,
\label{eq:ssm}
\end{equation}
where $\mathbf{A}_t, \mathbf{B}_t, \mathbf{W}_t, \mathbf{U}_t$ 
are input-dependent parameters generated by lightweight linear projections:
\begin{equation}
\mathbf{A}_t = \sigma(\mathbf{W}_a \mathbf{x}_t + \mathbf{b}_a), \quad
\mathbf{B}_t = \sigma(\mathbf{W}_b \mathbf{x}_t + \mathbf{b}_b),
\label{eq:gate}
\end{equation}
and $\sigma(\cdot)$ denotes the sigmoid activation, $\mathbf{A}_t$ acts as a decay gate controlling how much past information is preserved, 
and $\mathbf{B}_t$ modulates the injection of new input information.
Reshaping $\{\mathbf{y}_t\}_{t=1}^{T}$ back to the spatial layout yields the refined correlation embedding $\tilde{\mathbf{C}}\in \mathbb{R}^{H W\times N_c \times d_f}$.

For class aggregation, we adopt the same selective SSM formulation but process the category embedding 
$\tilde{\mathbf{C}}_{i,:} \in \mathbb{R}^{N_C\times d_f}$ at each spatial position sequentially. 
To maintain permutation consistency, classes are arranged in a fixed index order during training and inference. 

\noindent \textbf{Why is SSM-based Aggregation Better?}  
We summarize its advantages in two aspects:  
(1) Unlike attention, which explicitly mixes every token with every other token, the selective SSM aggregates information through a controlled recurrent state update. 
The decay gate $\mathbf{A}_t$ dynamically regulates how much of the previous state $\mathbf{h}_{t-1}$ should be preserved or forgotten. 
When correlations at step $t-1$ contain noise, a small decay value $\mathbf{A}_t \approx 0$ effectively forgets the unreliable state, preventing noisy patterns from propagating forward.
Meanwhile, the input gate $\mathbf{B}_t$ determines how strongly the current token $\mathbf{x}_t$ should influence the state, allowing clean and informative correlations to be injected when needed.
(2) The class-wise SSM refines inter-class dependencies in a more efficient manner 
and avoids the quadratic complexity $\mathcal{O}(N_C^2)$ inherent in attention.  
This linear-time formulation scales favorably when the open-vocabulary set is large, 
offering both higher efficiency (Table~\ref{Tab5_effic}) and better generalization to unseen categories.

\subsection{S$^2$-Corr} \label{sec:3.3}
Building upon the above enhanced baseline, 
we further propose a complete correlation refinement framework 
named S$^2$-Corr. 
It is designed to suppress domain-shift-induced noise from three perspectives: 
(1) modulation before aggregation, 
(2) a decay mechanism in the state-space model, 
and (3) a snake-shaped scanning strategy. 
Specifically, the first component injects informative cues into the correlation embeddings 
to guide the aggregation, 
while the latter two aim to suppress long-range noisy dependencies 
that arise during sequential propagation.

\noindent \textbf{Modulation Before Aggregation.}
Before performing spatial aggregation, 
we inject image-specific cues into the correlation embeddings 
$\mathbf{E}_{\pi(t), :}$ through a lightweight modulation step. 
Given the corresponding image features $\mathbf{F}_{\pi(t)}$, 
the correlation representation is adjusted as
\begin{equation} 
\hat{\mathbf{E}}_{\pi(t), :} =
\mathbf{E}_{\pi(t), :} \odot (1 + \gamma_{\pi(t)}) + \beta_{\pi(t)},
\end{equation}
where $(\gamma_{\pi(t)}, \beta_{\pi(t)})$ are modulation factors 
derived from $\mathbf{F}_{\pi(t)}$ via a linear projection. 
This operation injects image-dependent context 
and improves spatial consistency before the sequence aggregation.

Similarly, before class-wise aggregation, 
the category embeddings 
$\tilde{\mathbf{C}}_{i,:} \in \mathbb{R}^{N_C\times d_f}$ 
are refined using multi-domain textual prompts. 
We construct several domain-related templates such as 
``a photo of a cat at night'' or ``a photo of a cat in the rain'', 
which are encoded by a pretrained text encoder to obtain 
domain-aware text features $\mathbf{t}^{(d)}$. 
Each text feature produces a pair of modulation vectors 
$(\gamma^{(d)}, \beta^{(d)})$ to adjust the category embeddings:
\begin{equation} 
\hat{\mathbf{C}}_{i,:} =
\tilde{\mathbf{C}}_{i,:} \odot (1 + \gamma^{(d)}) + \beta^{(d)}.
\end{equation}
This text-driven modulation injects domain-specific semantics 
into the class representation, 
allowing the subsequent correlation fusion 
to better adapt to domain conditions.

\begin{figure}[!t]
    \begin{center}
    \centering \includegraphics[width=0.350\textwidth]{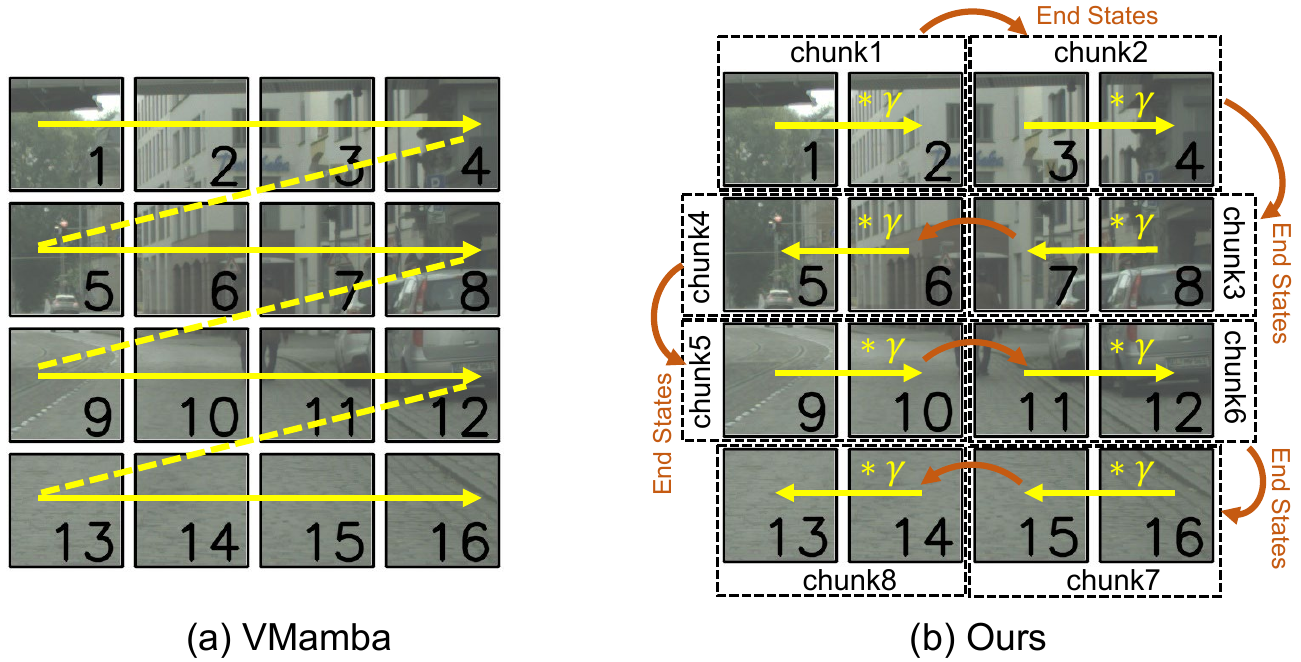}
    \end{center}
    \setlength{\abovecaptionskip}{-0.3 cm}
    \caption{Comparison of scanning strategies in state-space correlation aggregation between VMamba~\cite{liu2024vmambavisualstatespace} and ours. Our method introduces a learnable geometric decay ($*\gamma$) to suppress long-range noise within each chunk and uses a snake-shaped scanning strategy that preserves spatial continuity by passing end states between adjacent chunks.} 
    \label{Fig8_Mamba_comp}
\vspace{-0.3cm}
\end{figure}

\noindent\textbf{Learnable Geometric Decay Prior.} 
While the dynamic gate $\mathbf{A}_t$ in Eq.~(\ref{eq:gate}) enables data-dependent adaptation,
it may still carry the risk of propagating long-range noise (See Fig.~\ref{Fig7_corr_domain_shift} last column) under domain shifts.
To mitigate this, we introduce a learnable geometric decay prior,
\begin{equation}
\mathbf{A}_t^{\text{eff}} 
= \sigma(\mathbf{w})\,\sigma(\mathbf{W}_a \mathbf{x}_t + \mathbf{b}_a)
+ \bigl(1 - \sigma(\mathbf{w})\bigr)\,\boldsymbol{\gamma},
\label{eq:decay}
\end{equation}
where 
$\boldsymbol{\gamma}\!\in\!(0,1)^{K}$ is a head-specific 
spatial decay prior assigning each channel a base attenuation rate, 
and $\sigma(\mathbf{w})$ is a learnable coefficient balancing 
the data-driven gate and the geometric prior. 
The final update becomes
\begin{equation}
\mathbf{h}_t =
\mathbf{A}_t^{\text{eff}} \odot \mathbf{h}_{t-1}
+ \mathbf{B}_t \odot \mathbf{x}_t,
\label{eq:finalupdate}
\end{equation}
where $\odot$ denotes element-wise multiplication.
This design preserves a geometric attenuation pattern 
$\|\partial \mathbf{h}_t / \partial \mathbf{h}_{t-d}\| \!\propto\! (\mathbf{A}_t^{\text{eff}})^d$,
while allowing its decay rate to be learned from data, 
thus acting as a robust learnable geometric decay mechanism.

\begin{table*}[!t]
  \centering
   \renewcommand{\arraystretch}{1.05}
  \resizebox{0.995\textwidth}{!}{
\begin{tabular}{ccccccc|ccccc}
\toprule
\multirow{2}[2]{*}{Method} & \multirow{2}[2]{*}{Backbone} & \multirow{2}[2]{*}{Training Data} & \multicolumn{3}{c}{Dv-19} & \multirow{2}[2]{*}{Ave.} & \multicolumn{4}{c}{Dv-58}     & \multirow{2}[2]{*}{Ave.} \\
      &       &       & \textit{ACDC-19} & \textit{BDD-19} & \textit{Mapi-19} &       & \textit{ACDC-41} & \textit{BDD-41} & \textit{Mapi-30} & \textit{RW-10}&  \\
\midrule
CLIP-DINOiser(ECCV2024)~\cite{wysoczańska2024clipdinoiser} & \multicolumn{1}{c}{EVA02+DINOV2 ViT-L} &   Training-Free    & 25.3  & 28.5  & 27.7  & 27.2  & 31.9  & 34.0  & 26.9  & 21.6  & 28.6  \\
ClearCLIP(ECCV2024)~\cite{lan2024clearclip} & \multicolumn{1}{c}{EVA02 ViT-L} &    Training-Free   & 26.7  & 28.3  & 27.0  & 27.3  & 33.5  & 34.0  & 26.6  & 21.6  & 28.9  \\
ProxyCLIP (ECCV'24)~\cite{lan2024proxyclip} & \multicolumn{1}{c}{EVA02+SAM ViT-L} &  Training-Free     & 31.0  & 38.2  & 40.0  & 36.4  & 38.5  & 52.2  & 33.5  & 27.2  & 37.9  \\
\midrule
CAT-Seg (CVPR'24)~\cite{cho2024cat} & \multicolumn{1}{c}{EVA02 ViT-L/14} & COCO  & 30.3  & 31.9  & 36.1  & 32.8  & 33.2  & 42.8  & 28.6  & 27.6  & 33.1  \\
DeCLIP (CVPR'25)~\cite{wang2025declip} & \multicolumn{1}{c}{EVA02+DINOV2 ViT-L} & COCO  & 32.1  & 31.8  & 38.1  & 34.0  & 34.2  & 47.0  & 32.0  & 28.3  & 35.4  \\
\midrule
OVSeg (CVPR'23)~\cite{liang2023open} & \multicolumn{1}{c}{\multirow{10}[1]{*}{\shortstack{EVA02~\cite{eva02}\\ViT-B/16}}}
 & CS-7  & 36.6  & 40.9  & 45.8  & 41.1  & 41.2  & 48.3  & 31.7  & 32.6  & 38.5  \\
SAN (CVPR'23)~\cite{xu2023san} &       & CS-7  & 37.6  & 39.4  & 44.0  & 40.3  & 44.3  & 46.8  & 35.4  & 30.7  & 39.3  \\
CAT-Seg (CVPR'24)~\cite{cho2024cat} &       & CS-7  & 38.9  & 44.0  & 47.6  & 43.5  & 47.6  & 51.6  & 38.2  & \cellcolor[rgb]{ .886,  .937,  .855}36.5  & 43.5  \\
MAFT+(ECCV'24)~\cite{jiao2024collaborative} &       & CS-7  & 40.4  & 43.5  & \cellcolor[rgb]{ .886,  .937,  .855}50.2  & 44.7  & 46.4  & 48.8  & 38.1  & 35.4  & 42.2  \\
ESC-Net (CVPR'25)~\cite{ESCNet_2025_CVPR} &       & CS-7  & 40.4  & 43.6  & 45.9  & 43.3  & 47.3  & 49.2  & \cellcolor[rgb]{ .886,  .937,  .855}40.1  & 34.5  & 42.7  \\
MaskAdapter(CVPR'25)~\cite{Li_2025_CVPR}  &       & CS-7  & 38.9  & 47.7  & 49.9  & 45.5  & 48.3  & 51.2  & 39.8  & 36.1  & 43.8  \\
CLIPSelf (ICLR'24)~\cite{wu2023clipself} &       & CS-7+COCO & \cellcolor[rgb]{ .886,  .937,  .855}43.6  & 45.6  & 47.9  & 45.7  & \cellcolor[rgb]{ .886,  .937,  .855}49.4  & \cellcolor[rgb]{ .886,  .937,  .855}54.3  & 39.6  & \cellcolor[rgb]{ .886,  .937,  .855}36.5  & \cellcolor[rgb]{ .886,  .937,  .855}45.0  \\
RSC-CLIPSelf (ICLR'25)~\cite{qiu2025refining} &       & CS-7+COCO & 41.5  & 46.3  & 50.2  & \cellcolor[rgb]{ .886,  .937,  .855}46.0  & 50.0  & 53.0  & \cellcolor[rgb]{ .886,  .937,  .855}40.1  & 35.1  & 44.5  \\
CAT-Seg+AdvStyle~\cite{zhong2022adversarial} &       & CS-7  & 37.2  & 43.2  & 44.6  & 41.7  & 48.0  & 47.9  & 36.0  & 36.5  & 42.1  \\
CAT-Seg+DGInStyle~\cite{jia2024dginstyle} &       & CS-7  & 42.5  & \cellcolor[rgb]{ .886,  .937,  .855}49.8  & 49.4  & 44.6  & 46.1  & 53.7  & 36.8  & 36.0  & 43.2  \\
\rowcolor[rgb]{ .776,  .878,  .706} \textbf{S$^2$-Corr (Ours)} &       & CS-7  & \textbf{44.6} & \textbf{50.1} & \textbf{56.2} & \textbf{50.3} & \textbf{52.3} & \textbf{58.6} & \textbf{42.0} & \textbf{38.6} & \textbf{47.9} \\
\midrule
CAT-Seg (CVPR'24)~\cite{cho2024cat} & 
\multirow{4}{*}{\shortstack{EVA02\\ViT-L/14}} & 
CS-7  & 48.5  & 48.5  & 50.8  & 49.3  & 59.2  & 60.0  & 41.1  & 39.5  & 50.0  \\
MaskAdapter(CVPR'25)~\cite{Li_2025_CVPR}  &       & CS-7  & 48.1  & 49.4  & 54.6  & 50.7  & 59.7  & 59.7  & 41.8  & 36.1  & 49.3  \\
CLIPSelf (ICLR'24)~\cite{wu2023clipself} &       & CS-7+COCO & \cellcolor[rgb]{ .886,  .937,  .855}51.1  & \cellcolor[rgb]{ .886,  .937,  .855}53.0  & \cellcolor[rgb]{ .886,  .937,  .855}55.7  & \cellcolor[rgb]{ .886,  .937,  .855}53.3  & \cellcolor[rgb]{ .886,  .937,  .855}61.3  & 60.3  & \cellcolor[rgb]{ .886,  .937,  .855}44.7  & \cellcolor[rgb]{ .886,  .937,  .855}39.8  & \cellcolor[rgb]{ .886,  .937,  .855}51.5  \\
CAT-Seg+DGInStyle~\cite{jia2024dginstyle} &       & CS-7  & 50.4  & 51.7  & 50.4  & 50.8  & 60.2  & \cellcolor[rgb]{ .886,  .937,  .855}60.8  & 43.5  & 38.7  & 50.8  \\
\rowcolor[rgb]{ .776,  .878,  .706} \textbf{S$^2$-Corr (Ours)} &       & CS-7  & \textbf{54.3} & \textbf{53.1} & \textbf{60.0} & \textbf{55.8} & \textbf{62.0} & \textbf{61.7} & \textbf{47.4} & \textbf{41.9} & \textbf{53.2} \\
\bottomrule
\end{tabular}%

}
 	\setlength{\abovecaptionskip}{0.05 cm}
  \caption{Comparison of different OV-SS methods across different backbones under the \textit{Real-to-Real} OVDG-SS setting.  Dv-19 groups ACDC-19, BDD-19, and Mapi-19, while Dv-58 groups ACDC-41, BDD-41, Mapi-30, and RW-10.}
  \label{Tab2_ovdg_CS_vitL}%
  \vspace{-0.3cm}
\end{table*}%

\noindent \textbf{Chunk-wise Snake Scanning.}
To make the sequential update consistent with the 2D spatial layout, 
we divide the flattened sequence $\{\mathbf{x}_t\}_{t=1}^{T}$ into 
non-overlapping chunks of equal length $W$, 
each corresponding to a local region in the image grid.
Instead of scanning chunks strictly in a row-major order, 
which causes discontinuities at row boundaries, 
we adopt a snake-shaped traversal that alternates the scanning direction between consecutive rows. 
Within each chunk, the state is updated sequentially following Eq.~(\ref{eq:finalupdate}), 
and the final hidden state of the current chunk is passed to the next one,
$
\mathbf{h}^{\text{init}}_{k+1} \leftarrow \mathbf{h}^{\text{end}}_{k}.
$
This snake scanning strategy maintains spatial adjacency between rows 
and ensures smooth feature propagation across chunk boundaries. The illustration of our scanning strategy is shown in Fig.~\ref{Fig8_Mamba_comp}.

\section{Experiments}
\subsection{Dataset and Evaluation}
\noindent \textbf{Training Set.}
We use two datasets that have seven common driving categories (road, sidewalk, building, vegetation, sky, person, car) as source domains. 
\textit{CS-7} is constructed from Cityscapes~\cite{cordts2016cityscapes} by selecting these seven classes from the original 19 categories, yielding 2{,}975 real urban images.  
\textit{GTA-7} is the synthetic data from GTAV~\cite{richter2016GTA5}, containing 24{,}999 rendered images with the same label space.

\noindent \textbf{Testing Set.} We conduct evaluations on seven target domains.
\textit{\textit{ACDC-19}}~\cite{sakaridis2021acdc} contains 1{,}600 real images under extreme weather and follows the Cityscapes 19-class vocabulary.  
\textit{\textit{BDD-19}}~\cite{yu2020bdd100k} provides 1{,}000 images with diverse illumination and weather, also using the 19-class vocabulary.  
\textit{\textit{Mapi-19}} is from the Mapillary validation set~\cite{neuhold2017mapillary}, mapped to the same 19-class label space for evaluation. \textit{\textit{Mapi-30}} is built from the Mapillary training and validation sets~\cite{neuhold2017mapillary} by merging 65 categories into 30 coarse road-type classes, including bridges, tunnels, railways, and waterways (3{,}943 images). \textit{RW-10} comes from the ROADWork dataset~\cite{ghosh2025roadwork}, merged into 10 construction-related classes including workers, cones, work vehicles, and equipment (2{,}098 images). 

To introduce more open-vocabulary objects, we further construct \textit{\textit{ACDC-41}} and \textit{\textit{BDD-41}} using Stable Diffusion 2.1~\cite{rombach2022high} inpainting on ACDC and BDD images.  
The edited objects mimic additional entities commonly appearing on roads (e.g., daily objects, animals, obstacles). We manually filter distorted results and refine the segmentation labels, producing 1{,}000 high-quality synthetic images for each dataset with 41 categories.
(Details in Appendix). We refer \{ACDC-19, BDD-19 and Mapi-19\} as Dv-19 set which includes 19 classes while refer \{ACDC-41, BDD-41, Mapi-30 and RW-10\} as Dv-58 set which totally includes 58 classes.

\noindent \textbf{Task Construction.}
We study OVDG-SS under two evaluation paradigms: 
\textit{Synthetic-to-Real.}
We train on synthetic dataset \textit{GTA-7} with large gaps to real imagery, and evaluate on all real-world datasets (Dv-19 set and Dv-58 set).
\textit{Real-to-Real.}
We train on \textit{CS-7} and evaluate on the same target datasets to assess generalization from clean urban scenes to diverse real conditions. 

\begin{table*}[!t]
  \centering
   \renewcommand{\arraystretch}{1.05}
  \resizebox{0.995\textwidth}{!}{
\begin{tabular}{cccccccc|ccccc}
\toprule
\multirow{2}[2]{*}{Method} & \multirow{2}[2]{*}{Backbone} & \multirow{2}[2]{*}{Training Data} & \multicolumn{4}{c}{Dv-19}     & \multirow{2}[2]{*}{Ave.} & \multicolumn{4}{c}{Dv-58}     & \multirow{2}[2]{*}{Ave.} \\
      &       &       & \textit{CS-19} & \textit{ACDC-19} & \textit{BDD-19} & \textit{Mapi-19} &       & \textit{ACDC-41} & \textit{BDD-41} & \textit{Mapi-30} & \textit{RW-10} &  \\
\midrule
OVSeg (CVPR'23)~\cite{liang2023open} & \multirow{10}[1]{*}{\shortstack{EVA02 \\ ViT-B/16}} & GTA-7 & 41.2  & 34.9  & 42.5  & 45.1  & 40.9  & 45.1  & 52.6  & 35.5  & 32.9  & 41.5  \\
SAN (CVPR'23)~\cite{xu2023san} &       & GTA-7 & 43.3  & 36.7  & 42.0  & 48.1  & 42.5  & 47.7  & 49.0  & 37.8  & 30.3  & 41.2  \\
CAT-Seg (CVPR'24)~\cite{cho2024cat} &       & GTA-7 & 43.7  & 37.6  & 45.4  & 48.8  & 43.9  & \cellcolor[rgb]{ .886,  .937,  .855}51.2  & \cellcolor[rgb]{ .886,  .937,  .855}56.2  & 39.3  & \cellcolor[rgb]{ .886,  .937,  .855}35.7  & \cellcolor[rgb]{ .886,  .937,  .855}45.6  \\
MAFT+(ECCV'24)~\cite{jiao2024collaborative} &       & GTA-7 & 45.7  & 38.5  & 43.7  & 50.2  & 44.5  & 49.1  & 48.3  & 39.9  & 31.8  & 42.3  \\
ESC-Net (CVPR'25)~\cite{ESCNet_2025_CVPR} &       & GTA-7 & 42.2  & 35.9  & 44.6  & 48.1  & 42.7  & 49.4  & 52.4  & 37.9  & 33.3  & 43.3  \\
MaskAdapter(CVPR'25)~\cite{Li_2025_CVPR}  &       & GTA-7 & 46.0  & 36.4  & 45.7  & 50.1  & 44.6  & 50.1  & 49.2  & 38.1  & 32.7  & 42.5  \\
CLIPSelf (ICLR'24)~\cite{wu2023clipself} &       & GTA-7+COCO & \cellcolor[rgb]{ .886,  .937,  .855}48.1  & 37.6  & 47.9  & 51.1  & \cellcolor[rgb]{ .886,  .937,  .855}46.2  & 50.6  & 52.1  & \cellcolor[rgb]{ .886,  .937,  .855}40.4  & 34.6  & 44.4  \\
RSC-CLIPSelf (ICLR'25)~\cite{qiu2025refining} &       & GTA-7+COCO & 46.2  & 37.9  & 46.0  & \cellcolor[rgb]{ .886,  .937,  .855}51.6  & 45.4  & 51.1  & 50.7  & 40.3  & 32.0  & 43.5  \\
CAT-Seg+AdvStyle~\cite{zhong2022adversarial} &       & GTA-7 & 44.9  & 38.2  & 43.8  & 51.1  & 44.5  & 49.2  & 54.2  & 39.1  & 34.3  & 44.2  \\
CAT-Seg+DGInStyle~\cite{jia2024dginstyle} &       & GTA-7 & 44.4  & \cellcolor[rgb]{ .886,  .937,  .855}39.1  & \cellcolor[rgb]{ .886,  .937,  .855}48.1  & 51.4  & 45.8  & 50.7  & 55.9  & 39.6  & 34.0  & 45.1  \\
\rowcolor[rgb]{ .776,  .878,  .706} \textbf{S$^2$-Corr (Ours)} &       & GTA-7 & \textbf{49.3} & \textbf{40.4} & \textbf{48.6} & \textbf{54.5} & \textbf{48.2} & \textbf{51.3} & \textbf{56.4} & \textbf{42.4} & \textbf{36.7} & \textbf{46.7} \\
\midrule
CAT-Seg (CVPR'24)~\cite{cho2024cat} & 
\multirow{4}{*}{\shortstack{EVA02\\ViT-L/14}}      & GTA-7 & \cellcolor[rgb]{ .886,  .937,  .855}48.7  & 42.4  & 47.9  & 50.8  & 47.5  & 53.6  & 59.7  & \cellcolor[rgb]{ .886,  .937,  .855}41.1  & 38.2  & \cellcolor[rgb]{ .886,  .937,  .855}48.2  \\
MaskAdapter(CVPR'25)~\cite{Li_2025_CVPR}  &       & GTA-7 & 47.9  & 40.4  & 47.2  & 50.4  & 46.5  & 52.1  & 55.9  & 37.1  & 37.1  & 45.6  \\
CLIPSelf (ICLR'24)~\cite{wu2023clipself} &       & GTA-7+COCO & 47.8  & 43.1  & \cellcolor[rgb]{ .886,  .937,  .855}50.0  & 50.1  & \cellcolor[rgb]{ .886,  .937,  .855}47.8  & 51.2  & \cellcolor[rgb]{ .886,  .937,  .855}59.7  & 40.6  & \cellcolor[rgb]{ .886,  .937,  .855}39.7  & 48.0  \\
CAT-Seg+DGInStyle~\cite{jia2024dginstyle} &       & GTA-7 & 46.0  & \cellcolor[rgb]{ .886,  .937,  .855}43.9  & 48.9  & \cellcolor[rgb]{ .886,  .937,  .855}51.0  & 47.5  & \cellcolor[rgb]{ .886,  .937,  .855}53.7  & 59.0  & 37.9  & 38.0  & 47.2  \\
\rowcolor[rgb]{ .776,  .878,  .706} \textbf{S$^2$-Corr (Ours)} &       & GTA-7 & \textbf{52.2} & \textbf{44.2} & \textbf{50.3} & \textbf{53.3} & \textbf{49.9} & \textbf{55.4} & \textbf{59.8} & \textbf{41.5} & \textbf{40.8} & \textbf{49.4} \\
\bottomrule
\end{tabular}%
}
 	\setlength{\abovecaptionskip}{0.05 cm}
  \caption{Comparison of different OV-SS methods across various backbones under the \textit{synthetic-to-real} OVDG-SS setting.}
  \label{Tab3_ovdg_gta_vitL}%
  \vspace{-0.2cm}
\end{table*}%

\subsection{Implementation Details}
We implement our framework in Detectron2~\cite{wu2019detectron2}. 
The model is optimized using AdamW~\cite{loshchilov2017decoupled} with a learning rate of $2\times10^{-4}$ for the aggregation module and $2\times10^{-6}$ for the EVA-CLIP encoders. 
The correlation embedding dimension is set to 128, and the aggregation module uses 2 spatial blocks and 2 upsampling stages following CAT-Seg~\cite{cho2024cat}. 
The number of chunks is set to 16 and $\gamma$ is set as 0.8.
For vision encoders, the input resolution is $512$ for ViT-B/16 and $448$ for ViT-L/14, resulting in a $32{\times}32$ token grid for both.
In the visual encoder, only selected attention projection layers are updated, and in the text encoder, only projection weights within residual blocks are trainable.
Our model updates only 26M and 76.8M parameters using ViT-B/16 and ViT-L/14, respectively.
With a batch size of 4, we train the model by 20k iterations a single NVIDIA RTX~3090 GPU. 
\textbf{Notably, our method requires only about 2 hours training for EVA-CLIP ViT-B and 4 hours training for EVA-CLIP ViT-L} while achieving superior generalization on OVDG-SS.

\subsection{Comparison with State of The Art}
\noindent\textbf{Summary.} In Tables~\ref{Tab2_ovdg_CS_vitL} and~\ref{Tab3_ovdg_gta_vitL}, we compare our method with state-of-the-art training-free methods, open-vocabulary methods, and ``open-vocabulary + domain generalization~\cite{zhong2022adversarial}'' methods. Clearly, our method consistently achieves the best performance across all target datasets and backbones. 
Overall, OVDG exhibits strong generalization in both the Real-to-Real (CS-7 $\rightarrow$ Dv-19 / Dv-58) and Synthetic-to-Real (GTA-7 $\rightarrow$ Dv-19 / Dv-58) settings, achieving clear improvements over prior OV-SS approaches under both small- and large-vocabulary transfers.

\noindent \textbf{Results of Real-to-Real OVDG-SS.}
With ViT-B/16, S$^2$-Corr achieves 50.3\% on Dv-19, surpassing the previous best by 4.3 points and outperforming CAT-Seg and MaskAdapter by 6.8 and 4.8 points. 
On Dv-58, it reaches 47.9\%, improving the strongest prior result by 2.9 points with consistent gains across ACDC-41, BDD-41, and Mapi-30.
With ViT-L/14, S$^2$-Corr attains 55.8\% on Dv-19, improving the best existing result by 2.5 points and exceeding CAT-Seg and MaskAdapter by 6.5 and 5.1 points. 
On Dv-58, it reaches 53.2\%, outperforming the strongest baseline by 1.7 points while maintaining similarly stable improvements across all target domains.

\noindent \textbf{Results of Synthetic-to-Real OVDG-SS.} 
The GTA-to-real gap makes open-vocabulary transfer particularly challenging.
With ViT-B/16, on Dv-19, S$^2$-Corr achieves 48.2\%, outperforming the previous best (46.2\%) by 2.0 points and exceeding CAT-Seg and MaskAdapter by 4.3 and 3.6 points.
On Dv-58, it reaches 46.7\%, improving the strongest prior method by 1.1 points with consistent gains across ACDC-41, BDD-41, and Mapi-30.
With ViT-L/14, S$^2$-Corr attains 49.9\% on Dv-19, surpassing the previous best by 1.7 points and maintaining improvements across all real target domains.
On Dv-58, it reaches 49.4\%, exceeding the strongest baseline by 1.2 points while sustaining clear gains on ACDC-41, BDD-41, Mapi-30, and RW-10.

\begin{table}[t]
  \centering
   \renewcommand{\arraystretch}{1.05}
  \resizebox{0.495\textwidth}{!}{
\begin{tabular}{ccccccc}
\toprule
\multirow{2}[2]{*}{Note} & \multirow{2}[2]{*}{Methods} & \multicolumn{2}{c}{ViT-B/16} & \multicolumn{2}{c}{ViT-L/14} & \multirow{2}[2]{*}{Ave.} \\
      &       & Dv-19 & Dv-58 & Dv-19 & Dv-58 &  \\
\midrule
Base  & Cross-attention & 43.5  & 43.5  & 49.3  & 50.0  & 46.6  \\
\midrule
\#1   & Seletive SSM & 45.6  & 44.1  & 50.7  & 50.5  & 47.7  \\
\#2   & +Modulation  & 47.6  & 45.3  & 52.1  & 50.9  & 49.0  \\
\#3   & +Geometric decay & 48.3  & 46.4  & 53.2  & 51.8  & 49.9  \\
\#4   & +Chunk  & 49.6  & 47.3  & 55.3  & 52.7  & 51.2  \\
\#5   & +Snake scanning & \textbf{50.3} & \textbf{47.9} & \textbf{55.8} & \textbf{53.2} & \textbf{51.8} \\
\bottomrule
\end{tabular}%
}
 	\setlength{\abovecaptionskip}{0.05 cm}
  \caption{Ablation study of the proposed method under different design components on CS7 → Dv-19 and Dv-58 across backbones. Method \#1 replaces cross-attention by selective SSM. Methods of \#2–\#5 are progressively added on top of \#1.}
  \label{Tab4_alb}%
  \vspace{-0.2cm}
\end{table}%

\begin{figure*}[!t]
    \setlength{\abovecaptionskip}{-0.15cm}
    \begin{center}
    \centering 
    \includegraphics[width=0.99\textwidth ]{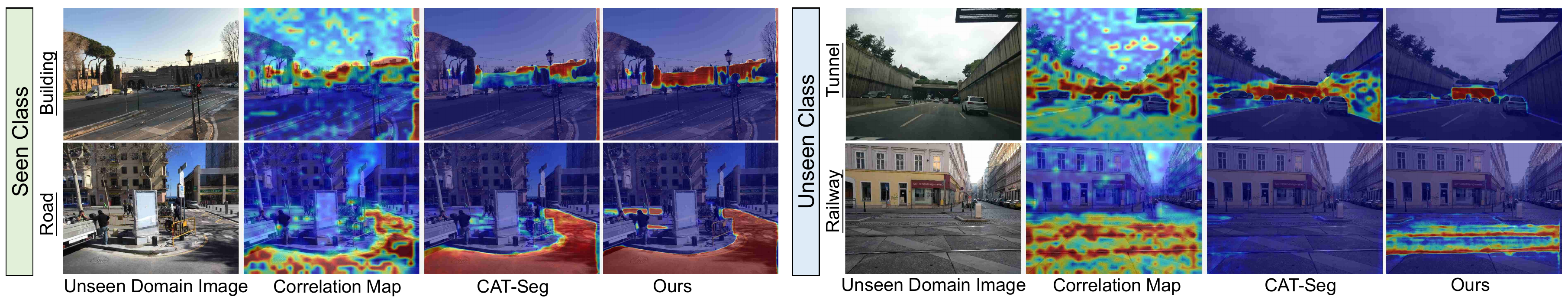} 
    \end{center}
    \caption{Comparison of text–image correlation aggregation on seen and unseen classes from unseen domains. Our method yields clearer and more localized text–image correlations than CAT-Seg~\cite{cho2024cat} and  Corelation Map (initially obtained by Eq.~\ref{initial_map}), improving both seen and unseen predictions under domain shifts.} 
    \setlength{\abovecaptionskip}{0.05 cm}
    \label{Fig4_Corr}
    \vspace{-0.3cm}
\end{figure*}

\begin{figure*}[!t]
    \setlength{\abovecaptionskip}{-0.15cm}
    \begin{center}
    \centering \includegraphics[width=0.98\textwidth]{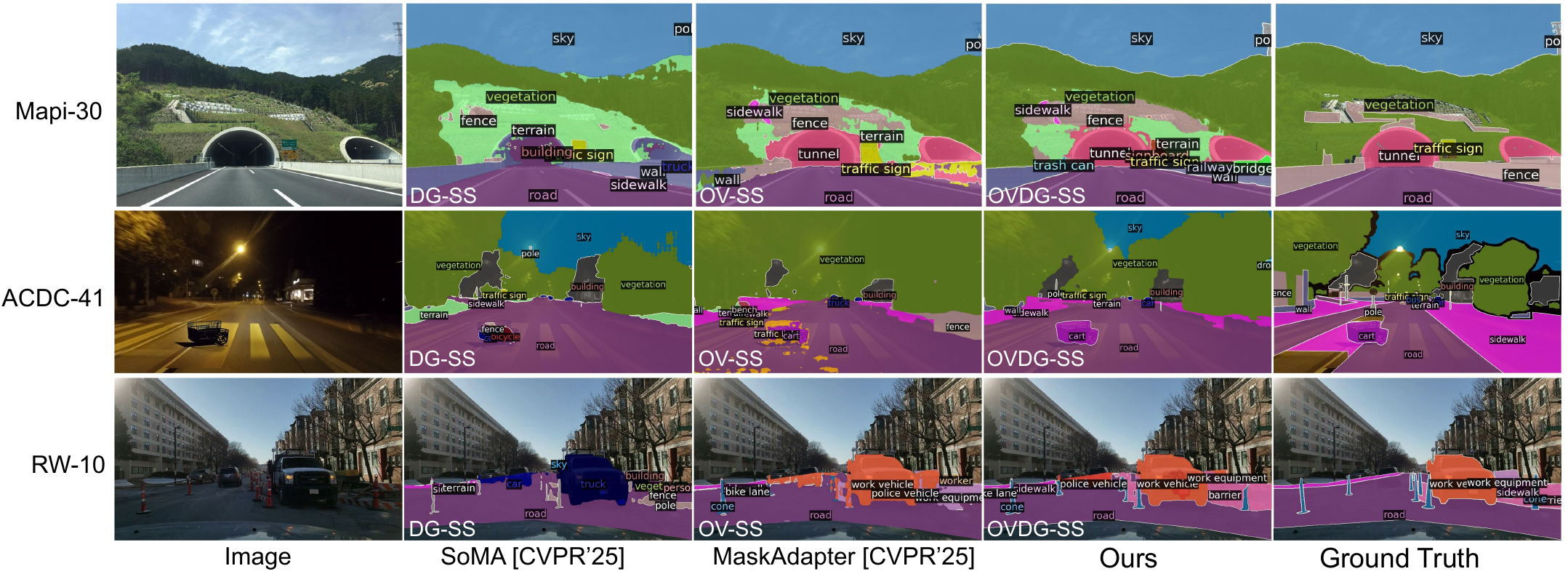} 
    \end{center}
    \setlength{\abovecaptionskip}{-0.25 cm}
    \caption{
    Qualitative comparisons on unseen domains under the OVDG-SS setting, with DG-SS method~\cite{yun2025soma} and OV-SS method~\cite{Li_2025_CVPR}. 
  } 
    \label{Fig5_seg}
    \vspace{-0.3cm}
\end{figure*}

\subsection{Ablation Study and Analysis}
\noindent \textbf{Ablation Study on Components.}
Table~\ref{Tab4_alb} reports ablations on CS-7~$\rightarrow$~Dv-19 and Dv-58 using both ViT-B/16 and ViT-L/14.
Three observations emerge.
\textit{First}, replacing the Swin-style cross-attention~\cite{cho2024cat} with our selective SSM baseline already yields clear gains (\#1 vs.\ Base), showing the advantage of sequential correlation aggregation over windowed attention. 
\textit{Second}, each enhancement further strengthens the baseline: modulation improves correlation modeling (\#2), and geometric decay brings additional gains by attenuating unreliable long-range interactions (\#3). 
\textit{Third}, the noise-suppression components—geometric decay and the chunk mechanism—deliver the largest improvements, particularly under the larger Dv-58 vocabulary. 
The final snake-scanning variant (\#5) achieves the best results across all settings, confirming the importance of controlling long-range propagation while preserving spatial continuity.
Overall, these ablations verify that each component contributes to a stronger correlation-refinement pipeline, together forming a robust solution for OVDG-SS.

\noindent \textbf{Correlation Aggregation Analysis.}
Fig.~\ref{Fig4_Corr} shows correlation refinement under domain shifts.
CAT-Seg often yields diffuse or misaligned responses, especially on unseen classes such as tunnel and railway.
S$^2$-Corr produces clearer and more localized correlations that better follow object structures, even in low-visibility scenes.
For seen and unseen classes alike, S$^2$-Corr suppresses noise, aligns with semantic boundaries, and ultimately improves OVDG-SS.

\noindent \textbf{Qualitative Results.}
Fig.~\ref{Fig5_seg} shows that DG-SS methods fail on unseen terrains, often misclassifying critical structures such as tunnels. 
OV-SS methods offer broader vocabulary but remain unstable under large domain shifts. In contrast, OVDG-SS delivers more accurate and consistent segmentation across diverse unseen domains, aligning well with true scene semantics.

\begin{table}[t]
  \centering
   \renewcommand{\arraystretch}{1.05}
  \resizebox{0.495\textwidth}{!}{
\begin{tabular}{cccccc}
\toprule
\multirow{2}[2]{*}{Model} & FPS   & FPS   & FPS   & GPU   & Training  \\
      & vocab19 &  vocab58 & vocab150 & Mem. (GB) & Time (min) \\
\midrule
CAT-Seg~\cite{cho2024cat} & 15.4  & 10.6  & 5.7   & 13.8  & 180 \\
ESC-Net ~\cite{ESCNet_2025_CVPR} & 15.0  & 9.9   & 5.1   & 15.7  & 220 \\
\midrule
Seletive SSM & 12.3  & 9.4   & 6.4   & 16.8  & 240 \\
\rowcolor[rgb]{ .886,  .937,  .855} \textbf{S$^2$-Corr (Ours)}  & \textbf{26.1} & \textbf{22.2} & \textbf{18.3} & \textbf{9.2} & \textbf{140} \\
\bottomrule
\end{tabular}%
}
 	\setlength{\abovecaptionskip}{0.05 cm}
  \caption{Inference and training efficiency of competing methods using a ViT-B/16 backbone. FPS is evaluated under different test vocabulary sizes. }
  \label{Tab5_effic}%
  \vspace{-0.4cm}
\end{table}%

\noindent\textbf{Efficiency Analysis.}
As shown in Table~\ref{Tab5_effic}, our method achieves clear efficiency advantages over existing correlation-based OV-SS models such as CAT-Seg and ESC-Net.
When the vocabulary size increases, the throughput of CAT-Seg and ESC-Net drops sharply to 5.7 and 5.1 FPS, whereas our model still maintains 18.3 FPS.
Our GPU memory usage is also significantly lower at 9.2 GB, compared with 13.8 GB and 15.7 GB, and the training time decreases from 180–220 minutes to 140 minutes.
These results demonstrate the superior computational efficiency and scalability of our design under large-vocabulary settings.
Although native SSMs scan all tokens sequentially and incur considerable overhead that can make them slower than CAT-Seg under large vocabularies, 
our chunk-wise aggregation preserves high parallelism and avoids unnecessary long-range interactions.
This leads to substantially faster inference and much lower memory consumption.
Additional efficiency comparisons are provided in Fig.~\ref{Fig2_Fps}.

\noindent \textbf{More detailed Analysis} on hyper-parameters, vocabulary sizes, training and benchmark are provided in the appendix.

\section{Conclusion}
\vspace{-0.2cm}
In this work, we introduce Open-Vocabulary Domain Generalization Semantic Segmentation (OVDG-SS) as a unified and realistic setting under both unseen domains and unseen categories.
Through a new autonomous-driving benchmark, we uncover a core weakness of VLM-based OV-SS models, \textit{i.e.,} their text–image correlations degrade sharply under domain shifts. To address this challenge, we propose S$^2$-Corr, a state-space correlation refinement mechanism that significantly improves cross-domain robustness while preserving open-vocabulary flexibility. We expect OVDG-SS and S$^2$-Corr to provide a practical step toward building more adaptable and reliable open-world segmentation systems for dynamic real-world environments.

\begin{table*}[!t]
  \centering
   \renewcommand{\arraystretch}{1.05}
  \resizebox{0.965\textwidth}{!}{
\begin{tabular}{ccccl}
\toprule
\textbf{dataset} & \textbf{From} & \textbf{Size} & \textbf{Resolution} & \multicolumn{1}{c}{\textbf{Classes}} \\
\midrule
CS-7  & Cityscapes Train~\cite{cordts2016cityscapes} & 2,975 & 2048×1024 & \multirow{2}[2]{*}{Rd, Sw, Bldg, Veg, Sky, Per, Car} \\
GTA-7 & GTA Train~\cite{richter2016GTA5} & 24999 & 2048×1024 &  \\
\midrule
ACDC-19 & ACDC Train~\cite{sakaridis2021acdc} & 1,600 & 1920×1080 & Rd, Sw, Bldg, Wal, Fen, Pol, TL, TS \\
BDD-19 & BDD100K Validation~\cite{yu2020bdd100k} & 1000  & 1280×720 & Veg, Ter, Sky, Per, Rid, Car, Trk, Bus, Trn, Mot, Bic \\
Mapi-19 & Mapillary Validation~\cite{yu2020bdd100k} & 2000  & 2k-4k  &  \\
\midrule
\multirow{2}[1]{*}{ACDC-41} & \multirow{2}[1]{*}{SD 2.1 + ACDC Train} & \multirow{2}[1]{*}{1000}  & \multirow{2}[1]{*}{1920×1080} & Rd, Sw, Bldg, Wal, Fen, Pol, TL, TS, Veg, Ter, Sky, Per,  \\
      &       &       &       & Rid, Car, Trk, Bus, Trn, Mot, Bic, Bag, Bal, Bar, Bnc, Brd,  \\
\multirow{2}[1]{*}{BDD-41} & \multirow{2}[1]{*}{SD 2.1+ BDD100K Validation} & \multirow{2}[1]{*}{1000} & \multirow{2}[1]{*}{1280×720} & Btl, Crt, Cat, Chr, Cow, Der, Dog, Drn, Elp, Hat, Hrs,  \\
      &       &       &       & Rbt, Shp, Tbl, Toy, Umb, Zbr \\
\midrule
\multirow{3}[2]{*}{Mapi-30} & \multirow{3}[2]{*}{Mapillary Training+Validation~\cite{yu2020bdd100k}} & \multirow{3}[2]{*}{3943} & \multirow{3}[2]{*}{2k-4k} & Rd, Sw, Bldg, Wal, Brg, Tun, TS, TL, Pol, Fen, Sky, \\
      &       &       &       &  Veg, Ter, Wtr, Snw, Snd, Per, Rid, Car, Trk, Bus, Trn, \\
      &       &       &       &  Bic, Mot, Anm, Sbd, Rwy, Bot, Chr, Tc \\
\midrule
RW-10 & ROADWork Training ~\cite{ghosh2025roadwork} & 2098  & 1280×720 & Rd, Sw, Bar, PV, WV, PO, Wkr, Con, ABd, TTC \\
\bottomrule
\end{tabular}%
}
 	\setlength{\abovecaptionskip}{0.05 cm}
  \caption{Summary of datasets used in the OVDG-SS benchmark.}
  \label{Tab1_detailed_dataset}%
  \vspace{-0.2cm}
\end{table*}%

\section{More Dataset Details}

As shown in Table~\ref{Tab1_detailed_dataset}, our OVDG-SS benchmark spans a progressively expanding set of semantic categories, beginning with the 7 basic driving classes from Cityscapes and GTA, including \textit{road, sidewalk, building, vegetation, sky, person, car}.

The 19-class datasets (ACDC-19~\cite{sakaridis2021acdc}, BDD-19~\cite{yu2020bdd100k}, Mapi-19~\cite{neuhold2017mapillary}) extend this space by incorporating additional urban background categories \textit{wall, fence, pole, traffic light, traffic sign, terrain} and dynamic traffic participants \textit{rider, truck, bus, train, motorcycle, bicycle}. These datasets differ in environmental conditions, covering adverse weather, varied illumination, and globally diverse street scenes.

The 41-class synthetic datasets (ACDC-41, BDD-41) further expand the vocabulary with a wide range of open-world objects introduced through our diffusion-based inpainting pipeline. In addition to the 19 urban classes, these datasets include the full set of inpainted objects across multiple semantic groups:  
animals (\textit{bird, cat, cow, deer, dog, elephant, horse, sheep, zebra}),  
man-made items (\textit{bag, ball, barrel, bench, bottle, cart, chair, hat, table, toy, umbrella}),  
machines and devices (\textit{drone, robot}),  
and other everyday objects.  
These inpainted categories enrich the open-vocabulary evaluation setting by introducing diverse unseen visual concepts that do not appear in the original driving datasets.

Mapillary-30 includes an extended set of fine-grained street-scene categories, covering structural elements \textit{bridge, tunnel}, natural entities \textit{water, snow, sand}, traffic-related elements \textit{traffic sign, traffic light, signboard, railway}, vehicles \textit{boat}, and everyday objects \textit{chair, trash can}, in addition to the core driving classes.

ROADWork-10~\cite{ghosh2025roadwork} focuses on construction-related semantics and introduces the unseen categories \textit{barrier, police vehicle, work vehicle, police officer, worker, cone, arrow board, TTC sign}.

This structured progression from basic driving categories to diverse inpainted open-world objects enables comprehensive evaluation of OVDG-SS models under increasingly complex and heterogeneous semantic spaces.

\section{More Implementation Details}

\subsection{Text prompt templates.}
In our implementation, we continue to generate text embeddings by composing natural-language sentences from class names, but we extend the original single-template design with a set of domain-aware prompts. Instead of relying solely on ``A photo of a \{class\}'', we experiment with ten prompt templates that describe variations in environment, lighting, weather, and scene context. These templates include sentences such as ``A photo of \{class\} in different environments'', ``An image of \{class\} under various conditions'', and ``A picture of \{class\} in rainy or foggy weather''. This experimental set of 10 domain-oriented prompts provides more diverse textual cues and encourages the text encoder to better capture cross-domain characteristics. A more systematic exploration of prompt learning is left for future work.

\subsection{Inference at High Resolutions}
To perform inference at high resolutions in urban-scene images (e.g., 2K) with ViT-B/16 and ViT-L/14, we adopt a sliding-window strategy that differs from the patch inference approach in~\cite{cho2024cat}.
Our implementation follows a window-based scanning scheme parameterized by \texttt{SW\_KERNEL}, \texttt{SW\_OVERLAP}, and \texttt{SW\_OUT\_RES}. Specifically, for ViT-L/14, we use a kernel size of 448 and an overlap ratio of 0.333, while the input image is first resized to an effective output resolution of 448$\times$896.
Given the resized image, we generate a grid of overlapping windows of size 448$\times$448. The stride is computed as $(1-\texttt{SW\_OVERLAP})$ times the kernel size, and when the spatial dimensions cannot be evenly covered, the window origin is backtracked to ensure full coverage of the image. Each window is independently normalized, fed through the dense CLIP encoder, and then passed into the segmentation head to produce local probability maps. These window-level predictions are then aggregated by summing the logits over all overlapping regions and normalizing by the accumulated weights, yielding a full-resolution prediction.

\subsection{Algorithm Details}
We provide the detailed algorithmic procedure of the proposed spatial aggregation in S$^2$-Corr module in Algorithm~\ref{alg:spatialmamba_mid}.

\begin{algorithm}[t]
\caption{Spatial Aggregation in S$^2$-Corr}
\label{alg:spatialmamba_mid}
\begin{algorithmic}[1]
\Require Correlation embeddings $\mathbf{E}\in\mathbb{R}^{C\times T\times H\times W}$, 
appearance guidance $\mathbf{G}$, Chunk size $L$, decay prior $\gamma$.
\Ensure Refined spatial embeddings $\widetilde{\mathbf{E}}$.

\State \textbf{Modulation before aggregation:}
compute modulation from $\mathbf{G}$ and update each location by Eq.~(6)
$
\widehat{\mathbf{E}}_{:,h,w}
= \mathbf{E}_{:,h,w}\odot(1+\gamma_{h,w})+\beta_{h,w}.
$

\State \textbf{Row-wise chunking:}
for each row, split the $W$ tokens into chunks of length $L$ and
initialize a row state $\mathbf{s}^{(0)}_{h}=\mathbf{0}$.

\State \textbf{Chunk-wise 1D state-space update:}
for each chunk and token $\mathbf{x}_t$, compute gates
$\alpha_t,\beta_t$ and the decayed update coefficient  
$
A_t=\sigma(w)\alpha_t + (1-\sigma(w))\gamma.
$
update the state and output token as
$
\mathbf{s}_t = A_t\odot\mathbf{s}_{t-1} + \beta_t\odot\mathbf{x}_t,\qquad
\mathbf{y}_t = W_t\mathbf{s}_t + U_t\mathbf{x}_t,
$
and store the final state of the chunk as $\mathbf{s}^{\text{end}}_{h,k}$.

\State \textbf{Cross-row state propagation:}
for each row $h>1$, initialize the new row state by
$
\mathbf{s}^{(0)}_{h} \leftarrow \eta_{\text{cross}}\,
\mathbf{s}^{\text{end}}_{h-1,k},
$
which induces snake-shaped spatial dependencies across rows.

\State \textbf{Reconstruction:}
reassemble the tokens $\mathbf{y}_t$ back to the $(H,W)$ layout to obtain
$\widetilde{\mathbf{E}}$.

\end{algorithmic}
\end{algorithm}

\begin{figure*}[!t]
    \begin{center}
    \centering \includegraphics[width=0.9850\textwidth]{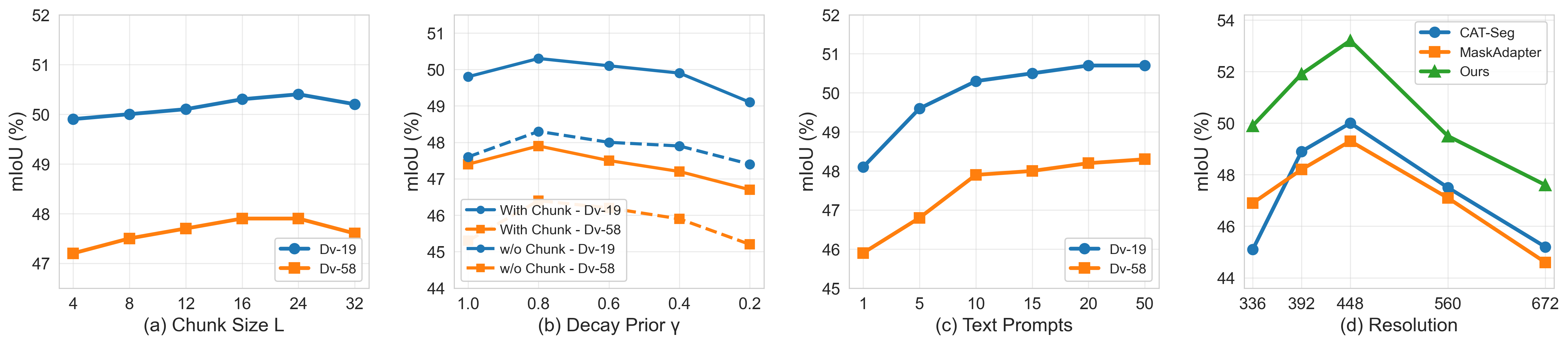}
    \end{center}
    \setlength{\abovecaptionskip}{-0.3 cm}
    \caption{Impact of hyperparameters: chunk size, prior decay, text prompt templates, and training resolution} 
    \label{Appendix_parameters}
\vspace{-0.1cm}
\end{figure*}

\section{More Results}

\subsection{Different VLM Pretrain}

\begin{table}[!h]
  \centering
   \renewcommand{\arraystretch}{1.05}
  \resizebox{0.4995\textwidth}{!}{
    \begin{tabular}{ccccccccc}
    \toprule
    VLM   & ACDC-19  & BDD-19  & Mapi-19 & ACDC-41  & BDD-41 &  Mapi-30 &  RW-10 & Ave. \\
    \midrule
    EVA-02-CLIP-L/14~\cite{eva02} & \underline{54.3} & \underline{53.1} & \underline{60} & \underline{62} & \underline{61.7} & 47.4  & 41.9  & 44.7  \\
    SigLIP-ViT-L/16~\cite{wang2024sclip} & 52.7  & 54.1  & 59.2  & 60.5  & 60.4  & \textbf{48.3} & \textbf{42.5} & 54.0  \\
    CLIP-ViT-L/14~\cite{clip} & 53.7  & 52.8  & 59.0    & 61.6  & 60.9  & 46.7  & 40.9  & 53.7  \\
    SelfCLIP-ViT-L/14~\cite{wu2023clipself} & \textbf{56.3} & \textbf{53.6} & \textbf{60.9} & \textbf{63.8} & \textbf{63.5} & \underline{48.1} & \underline{41.9} & \textbf{59.6} \\
    \bottomrule
    \end{tabular}%
 }
 	\setlength{\abovecaptionskip}{0.05 cm}
  \caption{Comparison of different Vision-Language Models (VLMs) under the OVDG-SS setting across multiple datasets. The best results are underlined.}
  \label{Tab8_diff_VLM}%
  \vspace{-0.2cm}
\end{table}%

As shown in Table~\ref{Tab8_diff_VLM}, among the evaluated VLMs, SelfCLIP-ViT-L/14 and SigLIP-ViT-L/16 achieve consistently stronger performance across the OVDG-SS datasets. 
Both models are post-distilled variants of CLIP that explicitly enhance the spatial alignment between image and text embeddings, making them inherently more suitable for pixel-level tasks such as semantic segmentation. 
In contrast, EVA-02-CLIP-L/14 and CLIP-ViT-L/14 do not include such spatial-consistency optimization. 
Although EVA-02 occasionally attains competitive scores due to its strong backbone pretraining, its performance varies more significantly across domains. 
Overall, these results indicate that VLMs with improved spatial alignment capabilities (SelfCLIP and SigLIP) provide more stable and robust open-vocabulary generalization, especially when transferring to unseen categories and diverse real-world conditions.

\subsection{Impact of Hyperparameters}

\noindent \textbf{Chunk size $L$.}  Fig.~\ref{Appendix_parameters}(a) reports the sensitivity of S$^2$-Corr to the chunk size $L$.  
The results show that performance remains remarkably stable across a wide range of $L$, indicating that the proposed chunked state-space design is inherently robust to this hyperparameter.  
Using a moderate chunk size (e.g., $L{=}12$--$16$) provides the best balance, while further increasing $L$ introduces longer-range dependencies that do not yield additional gains and may slightly degrade performance due to accumulated noise.  
Overall, the method shows weak sensitivity to $L$, highlighting the effectiveness of chunk-wise aggregation.

\noindent \textbf{Decay prior $\gamma$.}  Fig.~\ref{Appendix_parameters}(b) summarizes the effect of the decay prior $\gamma$ with and without the chunk mechanism. 
With chunking enabled, performance peaks around $\gamma{=}0.8$, while both overly large and overly small values lead to slight drops. 
This indicates that a moderate decay prior provides the right balance between retaining useful long-range information and suppressing noise accumulation. 
Without chunking, the model becomes more sensitive to $\gamma$, showing consistently lower performance and a larger performance gap between $\gamma{=}1$ and $\gamma{=}0.2$. 
These results demonstrate that chunking stabilizes the state-space recurrence and makes the decay prior significantly more robust.

\noindent \textbf{Number of text prompt templates.}  
We further examine the effect of varying the number of text prompt templates. 
As shown in Fig.~\ref{Appendix_parameters}(c), increasing the prompt pool from a single template to a more diverse set consistently boosts performance, confirming that multi-domain textual descriptions help the VLM produce more robust class embeddings for OVDG-SS. 
However, the performance gain saturates when the number exceeds $10$–$15$, while the computational cost continues to grow. 
We therefore adopt $10$ templates as a practical trade-off between accuracy and efficiency.

\noindent \textbf{Training Resolution.} We also investigate the impact of training resolution when fine-tuning EVA02-based ViT-L/14 models. 
As shown in Fig.~\ref{Appendix_parameters}(d), performance is highly sensitive to the choice of resolution: resolutions that deviate too much from the original pre-training size lead to significant degradation for all methods (CAT-Seg, MaskAdapter, and ours). 
Very small resolutions underfit the spatial details of downstream data (typically around $512\times1024$), whereas excessively large resolutions break the alignment with the ViT-L/14 pre-training scale, resulting in severe mismatch and loss of geometric consistency. 
Our method consistently achieves the best performance across settings, and peaks near the pre-training–aligned resolution (around $448$), confirming the importance of maintaining compatibility with the VFM’s original image scale during fine-tuning.

\subsection{Impact of Scanning ways in S$^2$-Corr}
\begin{table}[!h]
  \centering
  \renewcommand{\arraystretch}{1.05}
  \resizebox{0.4995\textwidth}{!}{
\begin{tabular}{cccccc}
\toprule
Variant & Row & Col & mIoU Dv-19 & mIoU Dv-58 & Training Time (Min) \\
\midrule
Row-only (ours) & $\checkmark$ & $\times$      & 50.3 & 47.9 & 140 \\
Col             & $\times$     & $\checkmark$  & 48.6 & 45.7 & 140 \\
Row+Col         & $\checkmark$ & $\checkmark$  & 50.4 & 48.1 & 165 \\
\bottomrule
\end{tabular}
}
\setlength{\abovecaptionskip}{0.05 cm}
\caption{Ablation on different scanning strategies used in the spatial aggregation module.}
\label{Tab9_Scanning_way}
\vspace{-0.2cm}
\end{table}

As shown in Table~\ref{Tab9_Scanning_way}, row-only scanning achieves a better balance between accuracy and efficiency. 
Pure column-wise scanning is clearly inferior on both domain settings, suggesting that horizontal spatial dependencies are more informative for correlation aggregation. 
The row+column variant brings only marginal improvements over row-only while introducing a notable increase in training time. 
Therefore, we adopt the row-only design as the default configuration in S$^2$-Corr.

\begin{table*}[!t]
  \centering
  \renewcommand{\arraystretch}{1.05}
  \resizebox{0.965\textwidth}{!}{
\begin{tabular}{cccccccccccc}
\toprule
\multicolumn{12}{c}{Training Data: GTA-19} \\
\midrule
      & Type  & Pretrain &
      {\cellcolor[rgb]{ .886,  .937,  .855}} CS-19 &
      {\cellcolor[rgb]{ .886,  .937,  .855}} BDD-19 &
      {\cellcolor[rgb]{ .886,  .937,  .855}} Mapi-19 &
      {\cellcolor[rgb]{ .886,  .937,  .855}} Ave. &
      {\cellcolor[rgb]{ .867,  .922,  .969}} Mapi-30 &
      {\cellcolor[rgb]{ .867,  .922,  .969}} RW-10 &
      {\cellcolor[rgb]{ .867,  .922,  .969}} ACDC-41 &
      {\cellcolor[rgb]{ .867,  .922,  .969}} BDD-41 &
      {\cellcolor[rgb]{ .867,  .922,  .969}} Ave. \\
\midrule
Rein (CVPR'24)~\cite{wei2024stronger} & DG    & EVA02-L & 65.3  & 60.5  & 64.9  & 63.6  & -       &   -    &    -   &   -    &  - \\
tqdm (ECCV'24)~\cite{pak2024textual} & DG    & EVA02-L & 68.9  & 59.2  & 70.1  & 66.1  & -       &   -    &    -   &   -    &  -  \\
SoMA (CVPR'25)~\cite{yun2025soma} & DG    & EVA02-L & 68.1  & 60.8  & 68.3  & 65.7  & -       &   -    &    -   &   -    &  -  \\
\midrule
MaskAdapter (CVPR'25)~\cite{Li_2025_CVPR} & OVSS  & EVA02-L & 61.5  & 55.5  & 65.7  & 60.9  & 48.5  & 41.0  & 63.3  & 61.8  & 53.7  \\
MAFT+(ECCV24)~\cite{jiao2024collaborative} & OVSS  & EVA02-L & 61.2  & 55.0  & 65.0  & 61.2  & 48.2  & 40.1  & 63.9  & 62.4  & 53.7  \\
Cat-Seg(CVPR'24) ~\cite{cho2024cat} & OVSS  & EVA02-L & 64.1  & 59.0  & 66.7  & 63.3  & 50.8  & 44.2  & 65.8  & 66.2  & 56.7  \\
\rowcolor[rgb]{ 1,  .949,  .8} 
\textbf{S$^2$-Corr (Ours)}  & OVDG  & EVA02-L & 65.2  & 61.8  & 67.6  & 64.9  & 51.9  & 44.8  & 66.2  & 66.5  & 57.4  \\
\bottomrule
\end{tabular}
}
\setlength{\abovecaptionskip}{0.05cm}
\caption{
Extended results under a larger 19-class training vocabulary. 
The main paper reports OVDG performance using a compact 7-class training set, 
this table presents results obtained with the full 19-class vocabulary (GTA-19). 
}
\label{Tab5_large_vocab}
\vspace{-0.2cm}
\end{table*}

\begin{figure*}[!t]
    \begin{center}
    \centering \includegraphics[width=0.9850\textwidth]{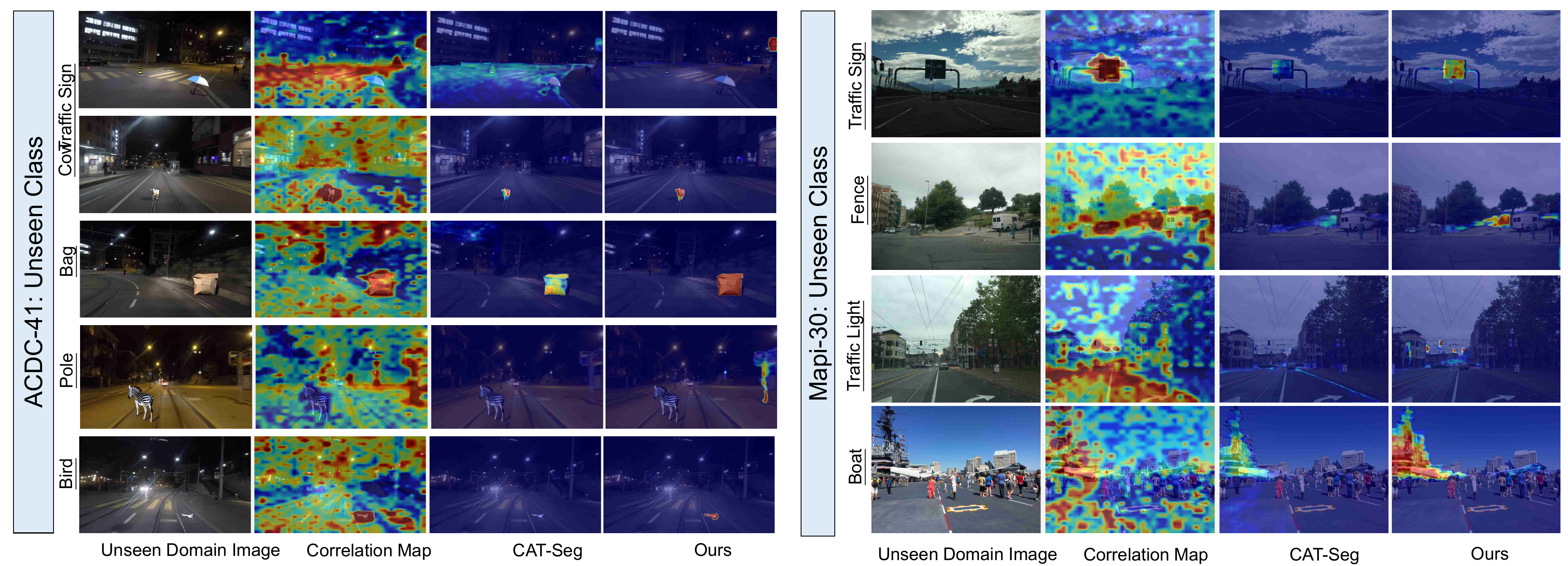}
    \end{center}
    \setlength{\abovecaptionskip}{-0.3 cm}
    \caption{More comparison of initial text–image correlation aggregation on seen and unseen classes from unseen domains.
}
    \label{Appendix_more_corr}
\vspace{-0.1cm}
\end{figure*}

\subsection{Training on Large Vocabulary}
Table~\ref{Tab5_large_vocab} shows that under the full 19-class GTA training vocabulary, our OVDG framework continues to demonstrate strong cross-domain generalization. 
Compared with proposal-based OVSS approaches such as MaskAdapter and MAFT+, correlation-refinement–based methods (e.g., Cat-Seg) show clear advantages on seen-class evaluation, confirming the effectiveness of refining text–image correlations rather than relying on region proposals alone. 
More importantly, when trained with the larger visible vocabulary, our OVDG method surpasses Cat-Seg on both seen-class benchmarks (CS-19, BDD-19, Mapi-19) and on unseen-class evaluations across Mapi-30, RW-10, ACDC-41, and BDD-41. 
Finally, although existing DG methods (Rein, tqdm, SoMA) can only evaluate seen classes, our OVDG achieves comparable performance on the same 19-class benchmarks while additionally providing unseen-class detection capability, which DG methods cannot offer.

\subsection{Detailed OVDG-SS results}
We provide additional experimental results using the ViT-L backbone in Table~\ref{Tab3_ovdg_gta_vitL_Details}.  
As shown in the table, our method consistently delivers highly competitive performance across all evaluated OVDG-SS settings.
As shown in Tables~\ref{Tab4_ovdg_per_class_19}, ~\ref{Tab5_ovdg_per_class_41}, ~\ref{Tab6_ovdg_per_class_30}, and~\ref{Tab7_ovdg_per_class_10}, we report the per-class OVDG-SS results on CS-7 for all evaluation datasets. The results demonstrate that our method achieves consistently strong performance on both seen and unseen classes.

\section{More Comparison}

\subsection{Correlation Refinement}

We further provide additional comparisons of the initial text-image correlation aggregation on unseen classes from unseen domains in Fig.~\ref{Appendix_more_corr}. 
Across multiple unseen categories and datasets, our method consistently filters domain-induced noise and produces more accurate spatial localization of unseen objects. 
These results reinforce the observations the main paper, demonstrating that our correlation refiner maintains robust semantic alignment even under significant domain shifts.

\subsection{Compared with Clost-Set DG}
Table~\ref{Tab5_large_vocab} further compares our OVDG framework with recent DG approaches under the full 19-class training vocabulary. Rein~\cite{wei2024stronger} improves domain robustness by introducing carefully engineered adapters that regularize the VLM’s vision encoder, yet its performance remains clearly lower than ours across all seen-class benchmarks. tqdm~\cite{pak2024textual} modifies the VLM through a different strategy, constructing a fixed set of category queries and effectively collapsing the open vocabulary space into a closed-set formulation. Despite this constrained design tailored for DG, our method still achieves competitive or superior results on CS-19, BDD-19, and Mapi-19. 
These comparisons show that although DG methods explicitly optimize VLMs for closed-set generalization, our OVDG approach achieves comparable or competitive performance on seen classes while also providing open-vocabulary recognition, a capability that DG methods do not possess.

\subsection{Compared with OOD Segmentation}
\begin{table*}[!h]
  \centering
   \renewcommand{\arraystretch}{1.05}
  \resizebox{0.895\textwidth}{!}{
\begin{tabular}{ccccccccccc}
\toprule
\multicolumn{1}{c|}{\multirow{2}[2]{*}{Methods}} & \multicolumn{2}{c|}{FS Static} & \multicolumn{2}{c|}{FS Lost\&Found} & \multicolumn{2}{c|}{SMIYC-Anomaly} & \multicolumn{2}{c|}{SMIYC-Obstacle} & \multicolumn{2}{c}{RoadAnomaly} \\
\multicolumn{1}{c|}{} & IoU   & \multicolumn{1}{c|}{mean F1} & IoU   & \multicolumn{1}{c|}{mean F1} & IoU   & \multicolumn{1}{c|}{mean F1} & IoU   & \multicolumn{1}{c|}{mean F1} & IoU   & mean F1 \\
\midrule
Synboost~\cite{Di_Biase_2021_CVPR} & 32.8  & 25.7  & 18.4  & 10.9  & 42.0  & 39.5  & 14.0  & 9.0   & 27.2  & 29.3  \\
PEBAL~\cite{esb_eccv_2022} & 26.9  & 13.3  & 6.4   & 2.6   & 42.4  & 35.1  & 6.7   & 1.1   & 33.8  & 23.9  \\
RPL+CoroCL~\cite{Liu_2023_ICCV} & 36.5  & 13.2  & 15.8  & 3.9   & 68.8  & 31.6  & 28.7  & 5.7   & 51.0  & 24.6  \\
S2M~\cite{Zhao_2024_CVPR_s2M} & 70.0  & 70.2  & 30.5  & 35.3  & 77.5  & 60.4  & 67.6  & 65.0  & 58.5  & 61.7  \\
\rowcolor[rgb]{ 1,  .949,  .8} Ours  & \textbf{74.6} & \textbf{76.7} & \textbf{36.3} & \textbf{42.2} & \textbf{82.9} & \textbf{69.7} & \textbf{76.2} & \textbf{73.8} & \textbf{66.5} & \textbf{66.9} \\
\bottomrule
\end{tabular}%
}
 	\setlength{\abovecaptionskip}{0.05 cm}
  \caption{
Comparison of open-set semantic segmentation on five anomaly benchmarks. 
Existing OOD segmentation methods (SynBoost, PEBAL, RPL+CoroCL, S2M) rely on COCO-Stuff segmentation supervision or additional anomaly masks for training. 
In contrast, our open-vocabulary segmentation model performs OOD detection purely through a text-extended vocabulary (Cityscapes 19 ID classes + 150 curated COCO-Stuff OOD concepts vocabulary) without using any COCO-Stuff pixel annotations. 
}
  \label{Tab10_openset_seg}%
  \vspace{-0.2cm}
\end{table*}%

\begin{figure*}[!t]
    \begin{center}
    \centering \includegraphics[width=0.9850\textwidth]{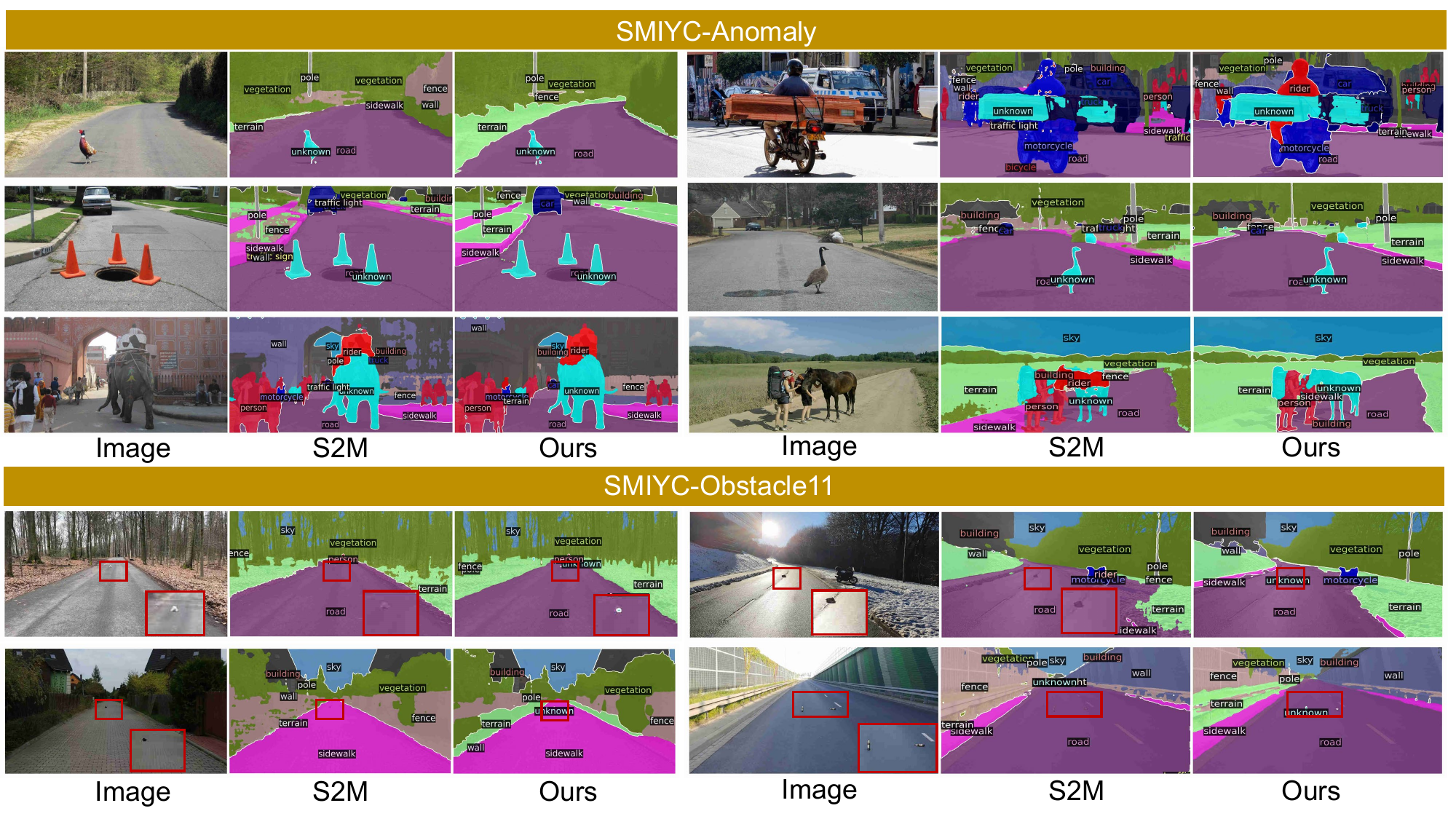}
    \end{center}
    \setlength{\abovecaptionskip}{-0.3 cm}
    \caption{
Qualitative comparison with S2M~\cite{Zhao_2024_CVPR_s2M} on SMIYC-Anomaly and SMIYC-Obstacle. 
}
    \label{Appendix_ood_seg}
\vspace{-0.1cm}
\end{figure*}

Unlike conventional out-of-domain(OOD) segmentation in urban-scene understanding~\cite{Di_Biase_2021_CVPR, esb_eccv_2022}, which focuses on detecting ``anything outside a fixed set of ID classes’’ and typically represents all unknown content with a single anomaly score or a binary OOD mask, 
our open-vocabulary segmentation framework addresses a more general and more semantically meaningful problem setting. 
In many real-world road-driving scenarios, merely flagging ``unknown'' is insufficient: different unknown objects (e.g., animals, construction tools, debris, or uncommon vehicles) require distinct semantics for safe decision-making. 
Open-vocabulary segmentation naturally generalizes the goal of OOD segmentation by enabling fine-grained recognition of arbitrary unseen concepts through text prompts, rather than collapsing them into a single ``unknown'' category. 
From this perspective, OOD segmentation can be viewed as a special case of open-vocabulary segmentation, and it is thus of interest to evaluate how our generalized formulation performs under standard OOD benchmarks.

To evaluate OOD localization, we leverage the fact that our model can operate on arbitrarily large textual vocabularies without architectural modification. 
For fair comparison with existing OOD segmentation methods, we construct a test-time vocabulary consisting of the 19 in-domain Cityscapes classes and 150 additional COCO-Stuff concepts. 
The COCO-Stuff categories are filtered to remove those semantically overlapping with Cityscapes classes, yielding a clean ID--OOD separation. 
During inference, the model predicts per-pixel labels over this enlarged vocabulary, and any pixel whose predicted label does not belong to the 19 Cityscapes ID classes is treated as OOD. 
Importantly, no COCO-Stuff pixel annotations or additional OOD-specific supervision are used; only the textual names of the 150 concepts are provided at test time.

To evaluate OOD localization, we leverage the fact that our model is an open-vocabulary segmentation framework that can directly operate on arbitrarily large textual label sets. 
For fair comparison with existing OOD segmentation methods, we construct a vocabulary consisting of the 19 in-domain Cityscapes classes and 150 additional COCO-Stuff concepts. 
The COCO-Stuff categories are filtered to remove those that overlap or closely align with Cityscapes classes, ensuring a clean semantic separation. 
During inference, the model predicts per-pixel labels over this enlarged vocabulary, and any pixel whose predicted label does not belong to the 19 Cityscapes ID classes is treated as OOD. 
No additional training or supervision from COCO-Stuff annotations is used; only the textual names of these 150 concepts are provided at test time.

Conventional OOD segmentation works commonly report threshold-based metrics such as FPR95, AUROC, or AUPR.  
However, these metrics depend heavily on calibrated OOD scores and threshold sweeps, making cross-method comparison unfair when models produce scores with different statistical ranges or when some methods (including ours) do not rely on explicit OOD scoring functions.  
To avoid such inconsistencies and to ensure a fair and model-agnostic evaluation, we follow the metric protocol of S2M~\cite{Zhao_2024_CVPR_s2M}, which directly evaluates the predicted OOD masks without requiring any score thresholding.
Following~\cite{Zhao_2024_CVPR_s2M}, we report both IoU and mean~F1 for OOD regions on five benchmarks: 
FS~Static~\cite{pinggera2016lost}, FS~Lost\&Found, SMIYC-Anomaly~\cite{chan2021segmentmeifyoucan}, SMIYC-Obstacle, and RoadAnomaly~\cite{lis2019detecting}. 
In contrast to existing OOD segmentation methods that rely on COCO-Stuff pixel annotations or additional anomaly-supervision signals, 
our model performs OOD detection \emph{solely} through text-driven vocabulary expansion, without using any OOD segmentation masks during training.

As shown in Table~\ref{Tab10_openset_seg}, across all datasets, our approach achieves the best overall performance, with notable improvements on SMIYC-Anomaly (+5.4 IoU over S2M) and SMIYC-Obstacle (+8.6 IoU). 
These results demonstrate that simple vocabulary expansion, enabled by open-vocabulary segmentation, provides a strong and scalable mechanism for OOD detection, outperforming specialized OOD detectors even without additional OOD supervision.

In Fig.~\ref{Appendix_ood_seg}, we further present qualitative comparisons with S2M on the SMIYC-Anomaly and SMIYC-Obstacle benchmarks. 
Our method produces cleaner OOD masks with sharper boundaries and better detection of small or thin obstacles. 
In cluttered or visually challenging scenes, S2M often misclassifies in-domain objects as unknown, whereas our open-vocabulary model preserves more stable predictions. 
These visual results align with the quantitative gains and demonstrate the improved robustness of our approach.

\subsection{More Visualization Comparisons}
\begin{figure*}[!t]
    \begin{center}
    \centering \includegraphics[width=0.9850\textwidth]{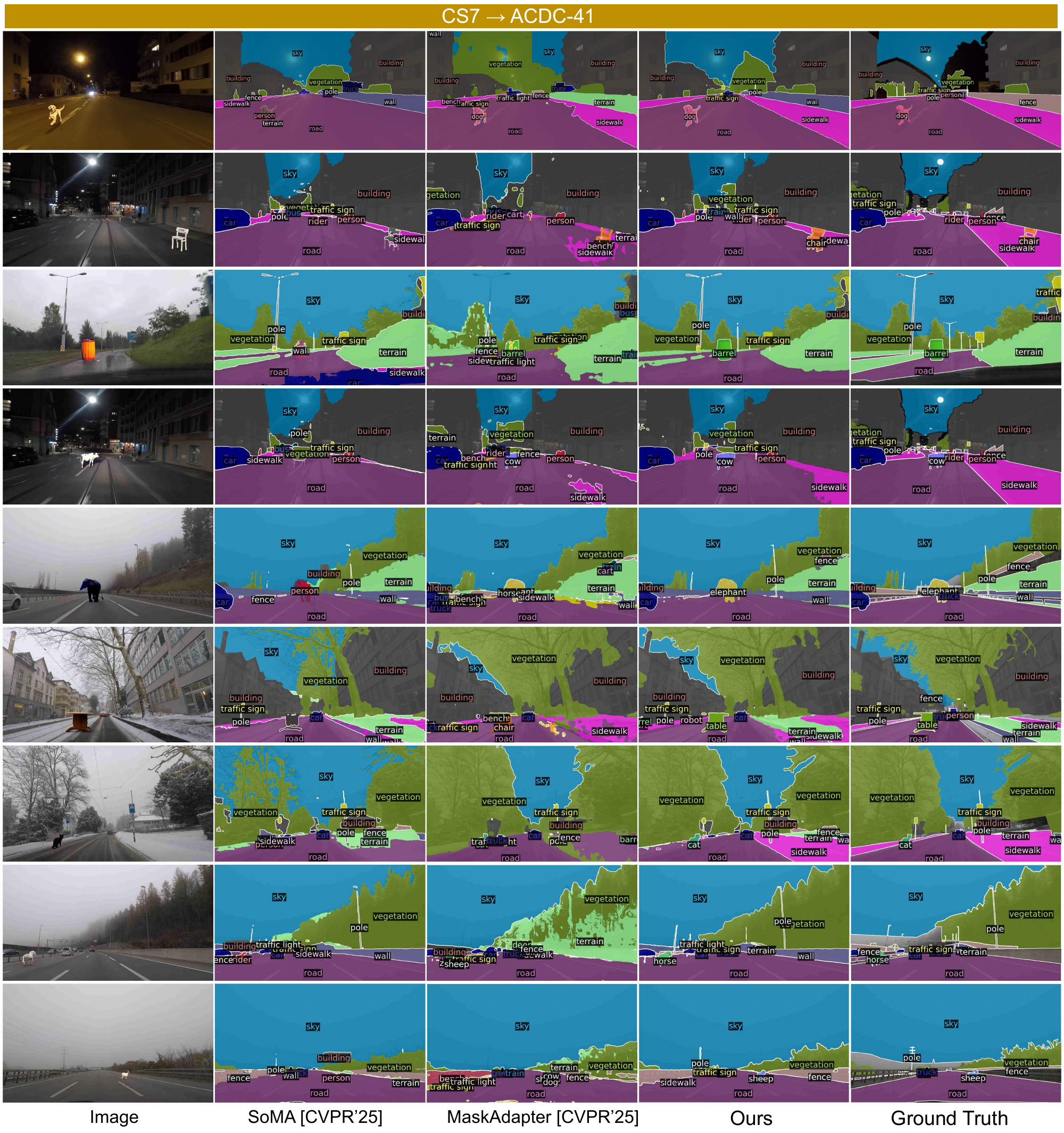}
    \end{center}
    \setlength{\abovecaptionskip}{-0.3 cm}
    \caption{More qualitative comparisons on unseen domains under the OVDG-SS setting on ACDC-41 dataset.
}
    \label{Appendix_seg_res1}
\vspace{-0.1cm}
\end{figure*}

\begin{figure*}[!t]
    \begin{center}
    \centering \includegraphics[width=0.9850\textwidth]{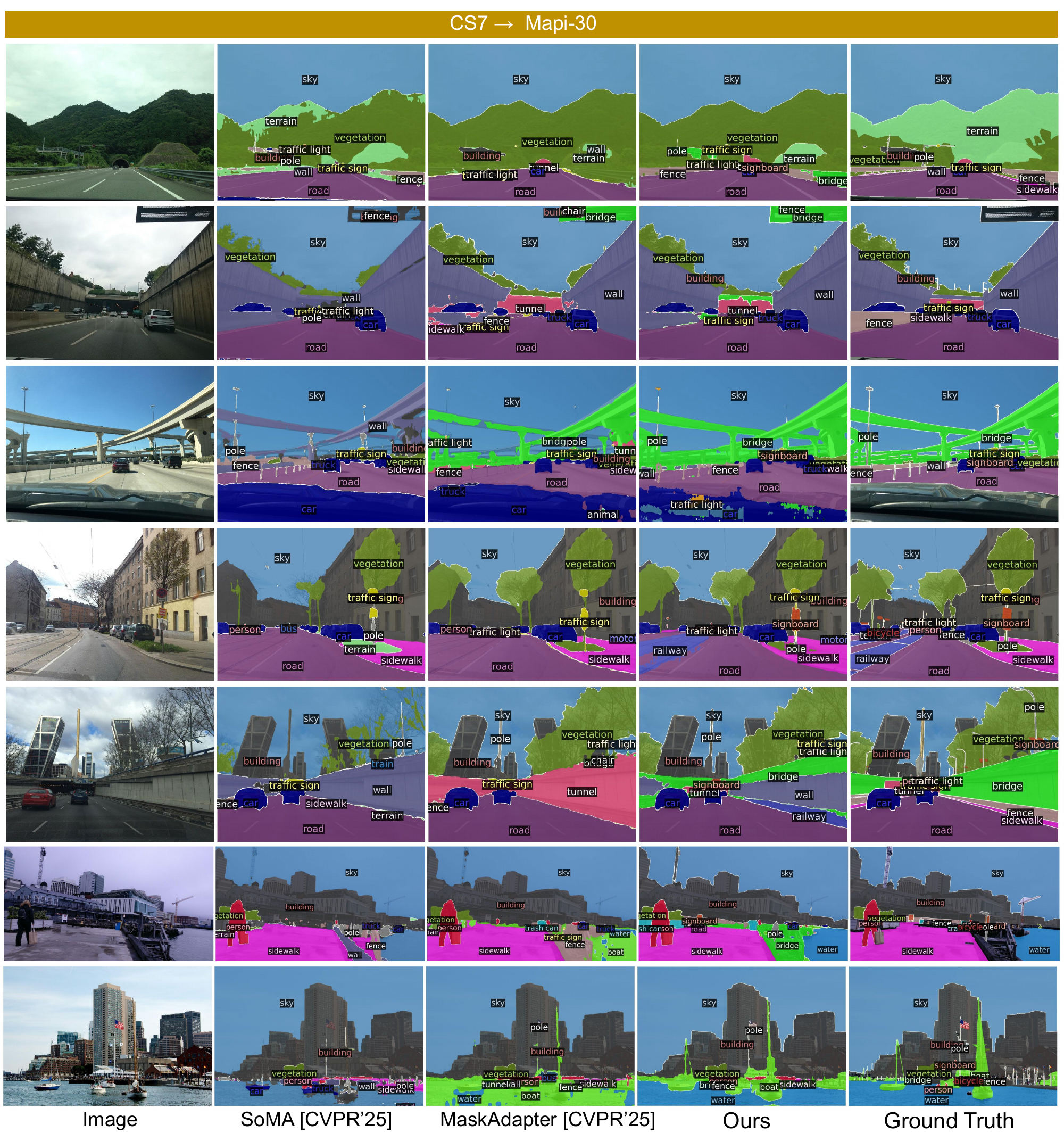}
    \end{center}
    \setlength{\abovecaptionskip}{-0.3 cm}
    \caption{More qualitative comparisons on unseen domains under the OVDG-SS setting on Mapi-30 datasets.
}
    \label{Appendix_seg_res2}
\vspace{-0.1cm}
\end{figure*}

\begin{figure*}[!t]
    \begin{center}
    \centering \includegraphics[width=0.9850\textwidth]{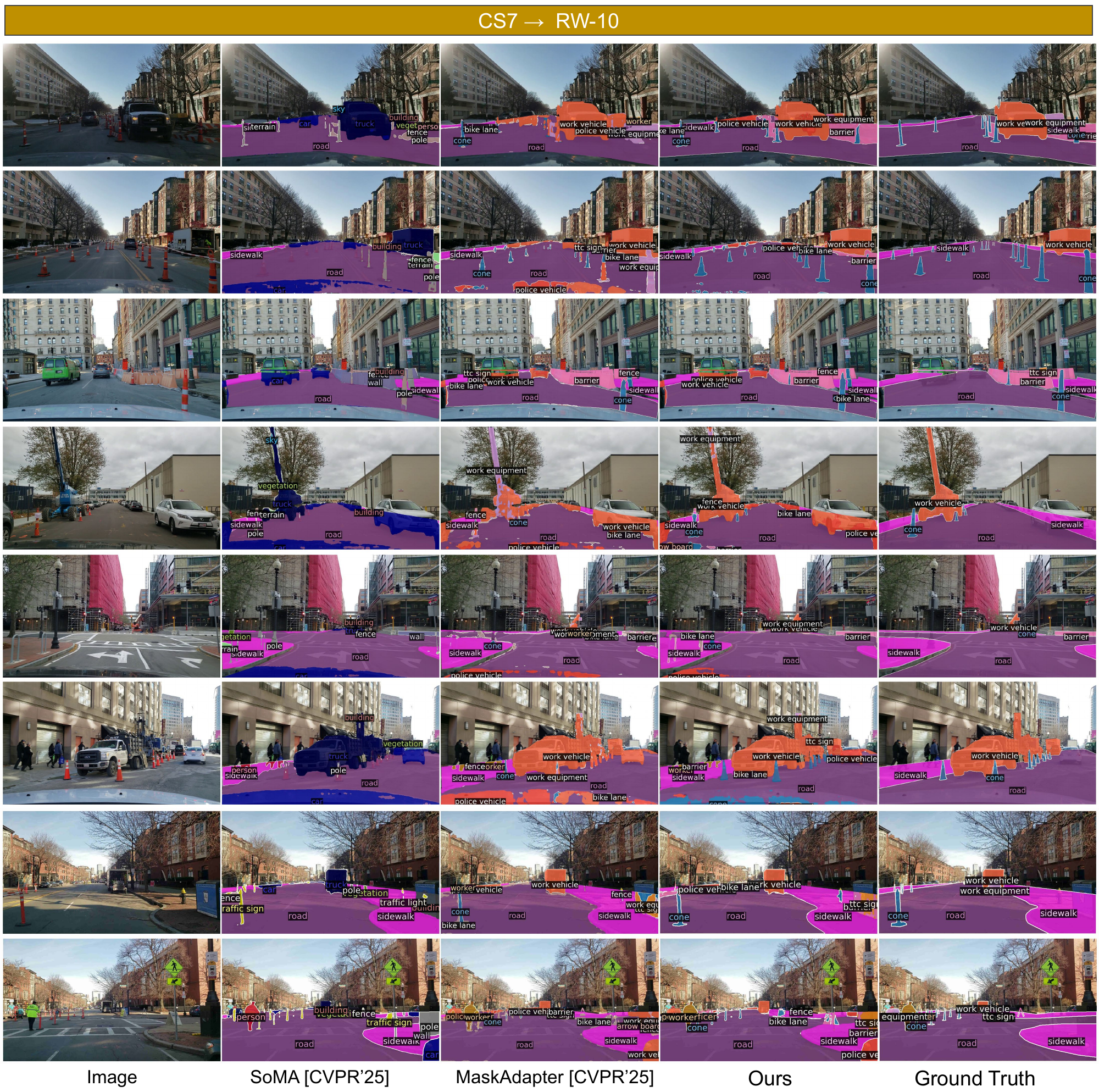}
    \end{center}
    \setlength{\abovecaptionskip}{-0.3 cm}
    \caption{More qualitative comparisons on unseen domains under the OVDG-SS setting on RW-10 datasets.
}
    \label{Appendix_seg_res3}
\vspace{-0.1cm}
\end{figure*}

As shown in Fig.~\ref{Appendix_seg_res1}, Fig.~\ref{Appendix_seg_res2}, and Fig.~\ref{Appendix_seg_res3}, we provide additional visualization comparisons on the CS7 $\rightarrow$ ACDC-41, Mapi-30, and RW-10 benchmarks. These results further verify that our method produces cleaner boundaries, reduces domain-induced noise, and achieves more consistent predictions across diverse unseen domains.

\section{Limitations and Outlook}
While we introduce the OVDG task setting for the first time and propose a novel enhancement method to mitigate domain-shift challenges in open-vocabulary semantic segmentation, several limitations remain. 
First, although our correlation refinement strategy effectively suppresses domain-induced noise, more efficient or principled mechanisms may further improve robustness, especially under severe distribution shifts. 
Second, our approach primarily refines correlations at inference time; strengthening the underlying VLM features to be intrinsically resistant to domain factors is an important direction for future work. 
Third, the current OVDG benchmark provides only a moderate set of open-vocabulary classes and test scenarios. 
We plan to substantially expand the benchmark with richer and more diverse categories, particularly those critical for driving safety, to better evaluate open-vocabulary generalization in real-world environments.

\begin{table*}[!t]
  \centering
   \renewcommand{\arraystretch}{1.05}
  \resizebox{0.995\textwidth}{!}{
\begin{tabular}{ccccccc|ccccc}
\toprule
\multirow{2}[2]{*}{Method} & \multirow{2}[2]{*}{Backbone} & \multirow{2}[2]{*}{Training Data} & \multicolumn{3}{c}{Dv-19} & \multirow{2}[2]{*}{Ave.} & \multicolumn{4}{c}{Dv-58}     & \multirow{2}[2]{*}{Ave.} \\
      &       &       & ACDC-19 & BDD-19 & Mapi-19 &       & ACDC-41 & BDD-41 & Mapi-30 & RW-10 &  \\
\midrule
CLIP-DINOiser(ECCV2024)~\cite{wysoczańska2024clipdinoiser} & CLIP+DinoV2 ViT-L/14 & Training-Free  & 25.3  & 28.5  & 27.7  & 27.2  & 31.9  & 34.0  & 26.9  & 21.6  & 28.6  \\
ClearCLIP(ECCV2024)~\cite{lan2024clearclip} & CLIP ViT-L/14 & Training-Free  & 26.7  & 28.3  & 27.0  & 27.3  & 33.5  & 34.0  & 26.6  & 21.6  & 28.9  \\
ProxyCLIP (ECCV’24) & CLIP+SAM ViT-L/14 & Training-Free  & 31.0  & 38.2  & 40.0  & \multicolumn{1}{c}{36.4} & 38.5  & 52.2  & 33.5  & 27.2  & 37.9  \\
\midrule
OVSeg (CVPR'23)~\cite{liang2023open} & \multirow{10}[1]{*}{EVA02 
ViT-L/14} & CS-7  & 41.9  & 42.6  & 43.9  & 42.8  & 51.2  & 49.3  & 33.7  & 35.7  & 42.5  \\
SAN (CVPR'23)~\cite{xu2023san} &       & CS-7  & 46.6  & 47.0  & 48.3  & 47.3  & 57.5  & 56.9  & 38.0  & 34.8  & 46.8  \\
CAT-Seg (CVPR'24)~\cite{cho2024cat} &       & CS-7  & 48.5  & 48.5  & 50.8  & 49.3  & 59.2  & 60.0  & 41.1  & 39.5  & 50.0  \\
\bottomrule
\end{tabular}%
}
 	\setlength{\abovecaptionskip}{0.05 cm}
  \caption{Comparison of different OV-SS methods using ViT-L/14 backbone  under OVDG-SS setting.}
  \label{Tab3_ovdg_gta_vitL_Details}%
  \vspace{-0.2cm}
\end{table*}%

\begin{table*}[!t]
  \centering
   \renewcommand{\arraystretch}{1.05}
  \resizebox{0.995\textwidth}{!}{
\begin{tabular}{rrrrrrrrrrrrrrrrrrrrr}
      &       &       &       &       &       &       &       &       &       &       &       &       &       &       &       &       &       &       &       &  \\
\midrule
\multicolumn{21}{c}{\textbf{CS-7→ACDC-19 ViT-B/16}} \\
\midrule
\multicolumn{1}{c}{Methods} & \multicolumn{1}{c}{\cellcolor[rgb]{ .886,  .937,  .855}Rd} & \multicolumn{1}{c}{\cellcolor[rgb]{ .886,  .937,  .855} Sw} & \multicolumn{1}{c}{\cellcolor[rgb]{ .886,  .937,  .855} Bldg} & \multicolumn{1}{c}{\cellcolor[rgb]{ .886,  .937,  .855} Veg} & \multicolumn{1}{c}{\cellcolor[rgb]{ .886,  .937,  .855} Sky} & \multicolumn{1}{c}{\cellcolor[rgb]{ .886,  .937,  .855} Per} & \multicolumn{1}{c}{\cellcolor[rgb]{ .886,  .937,  .855} Car} & \multicolumn{1}{c}{\cellcolor[rgb]{ .867,  .922,  .969}  Wal} & \multicolumn{1}{c}{\cellcolor[rgb]{ .867,  .922,  .969}  Fen} & \multicolumn{1}{c}{\cellcolor[rgb]{ .867,  .922,  .969}  Pol} & \multicolumn{1}{c}{\cellcolor[rgb]{ .867,  .922,  .969}  TL} & \multicolumn{1}{c}{\cellcolor[rgb]{ .867,  .922,  .969}  TS} & \multicolumn{1}{c}{\cellcolor[rgb]{ .867,  .922,  .969}  Ter} & \multicolumn{1}{c}{\cellcolor[rgb]{ .867,  .922,  .969}  Rid} & \multicolumn{1}{c}{\cellcolor[rgb]{ .867,  .922,  .969}  Trk} & \multicolumn{1}{c}{\cellcolor[rgb]{ .867,  .922,  .969}  Bus} & \multicolumn{1}{c}{\cellcolor[rgb]{ .867,  .922,  .969}  Trn} & \multicolumn{1}{c}{\cellcolor[rgb]{ .867,  .922,  .969}  Mot} & \multicolumn{1}{c|}{\cellcolor[rgb]{ .867,  .922,  .969}  Bic} & \multicolumn{1}{c}{\cellcolor[rgb]{ .867,  .922,  .969} mIoU} \\
\end{tabular}%
}
 	\setlength{\abovecaptionskip}{0.00 cm}
  \caption{Per-class results for OVDG-SS. 
Green entries denote seen classes, while blue entries correspond to unseen classes.}
  \label{Tab4_ovdg_per_class_19}%
  \vspace{-0.2cm}
\end{table*}%

\begin{table*}[!t]
  \centering
   \renewcommand{\arraystretch}{1.05}
  \resizebox{0.995\textwidth}{!}{
\begin{tabular}{ccccccccccccccccccccccccccccccccccccccccccc}
\toprule
\multicolumn{43}{c}{\textbf{CS-7→ACDC-41 ViT-B/16}} \\
\midrule
      & \cellcolor[rgb]{ .886,  .937,  .855}Rd & \cellcolor[rgb]{ .886,  .937,  .855} Sw & \cellcolor[rgb]{ .886,  .937,  .855} Bldg & \cellcolor[rgb]{ .886,  .937,  .855} Veg & \cellcolor[rgb]{ .886,  .937,  .855} Sky & \cellcolor[rgb]{ .886,  .937,  .855} Per & \cellcolor[rgb]{ .886,  .937,  .855} Car & \cellcolor[rgb]{ .867,  .922,  .969}  Wal & \cellcolor[rgb]{ .867,  .922,  .969}  Fen & \cellcolor[rgb]{ .867,  .922,  .969}  Pol & \cellcolor[rgb]{ .867,  .922,  .969}  TL & \cellcolor[rgb]{ .867,  .922,  .969}  TS & \cellcolor[rgb]{ .867,  .922,  .969}  Ter & \cellcolor[rgb]{ .867,  .922,  .969}  Rid & \cellcolor[rgb]{ .867,  .922,  .969}  Trk & \cellcolor[rgb]{ .867,  .922,  .969}  Bus & \cellcolor[rgb]{ .867,  .922,  .969}  Trn & \cellcolor[rgb]{ .867,  .922,  .969}  Mot & \cellcolor[rgb]{ .867,  .922,  .969}  Bic & \cellcolor[rgb]{ .867,  .922,  .969} Bag & \cellcolor[rgb]{ .867,  .922,  .969}  Bal & \cellcolor[rgb]{ .867,  .922,  .969}  Bar & \cellcolor[rgb]{ .867,  .922,  .969}  Bnc & \cellcolor[rgb]{ .867,  .922,  .969}  Brd & \cellcolor[rgb]{ .867,  .922,  .969}  Btl & \cellcolor[rgb]{ .867,  .922,  .969}  Crt & \cellcolor[rgb]{ .867,  .922,  .969}  Cat & \cellcolor[rgb]{ .867,  .922,  .969}  Chr & \cellcolor[rgb]{ .867,  .922,  .969}  Cow & \cellcolor[rgb]{ .867,  .922,  .969}  Der & \cellcolor[rgb]{ .867,  .922,  .969}  Dog & \cellcolor[rgb]{ .867,  .922,  .969}  Drn & \cellcolor[rgb]{ .867,  .922,  .969}  Elp & \cellcolor[rgb]{ .867,  .922,  .969}  Hat & \cellcolor[rgb]{ .867,  .922,  .969}  Hrs & \cellcolor[rgb]{ .867,  .922,  .969}  Rbt & \cellcolor[rgb]{ .867,  .922,  .969}  Shp & \cellcolor[rgb]{ .867,  .922,  .969} Tbl & \cellcolor[rgb]{ .867,  .922,  .969}  Toy & \cellcolor[rgb]{ .867,  .922,  .969}  Umb & \cellcolor[rgb]{ .867,  .922,  .969}  Zbr & \cellcolor[rgb]{ .867,  .922,  .969} mIoU \\
\bottomrule
\end{tabular}%
}
 	\setlength{\abovecaptionskip}{0.00 cm}
  \caption{Per-class results for OVDG-SS in CS-7→ACDC-41 and BDD-41.}
  \label{Tab5_ovdg_per_class_41}%
  \vspace{-0.2cm}
\end{table*}%

\begin{table*}[!t]
  \centering
   \renewcommand{\arraystretch}{1.05}
  \resizebox{0.995\textwidth}{!}{
\begin{tabular}{cccccccccccccccccccccccccccccccc}
\toprule
\multicolumn{32}{c}{\textbf{CS-7→Map-30 ViT-B/16}} \\
\midrule
      & \cellcolor[rgb]{ .886,  .937,  .855}Rd & \cellcolor[rgb]{ .886,  .937,  .855} Sw & \cellcolor[rgb]{ .886,  .937,  .855} Bldg & \cellcolor[rgb]{ .886,  .937,  .855} Veg & \cellcolor[rgb]{ .886,  .937,  .855} Sky & \cellcolor[rgb]{ .886,  .937,  .855} Per & \cellcolor[rgb]{ .886,  .937,  .855} Car & \cellcolor[rgb]{ .867,  .922,  .969}  Wal & \cellcolor[rgb]{ .867,  .922,  .969}  Fen & \cellcolor[rgb]{ .867,  .922,  .969}  Pol & \cellcolor[rgb]{ .867,  .922,  .969}  TL & \cellcolor[rgb]{ .867,  .922,  .969}  TS & \cellcolor[rgb]{ .867,  .922,  .969}  Ter & \cellcolor[rgb]{ .867,  .922,  .969}  Rid & \cellcolor[rgb]{ .867,  .922,  .969}  Trk & \cellcolor[rgb]{ .867,  .922,  .969}  Bus & \cellcolor[rgb]{ .867,  .922,  .969}  Trn & \cellcolor[rgb]{ .867,  .922,  .969}  Mot & \cellcolor[rgb]{ .867,  .922,  .969}  Bic & \cellcolor[rgb]{ .867,  .922,  .969} Chr & \cellcolor[rgb]{ .867,  .922,  .969}  Brg & \cellcolor[rgb]{ .867,  .922,  .969}  Tun & \cellcolor[rgb]{ .867,  .922,  .969}  Wtr & \cellcolor[rgb]{ .867,  .922,  .969}  Snw & \cellcolor[rgb]{ .867,  .922,  .969}  Snd & \cellcolor[rgb]{ .867,  .922,  .969}  Anm & \cellcolor[rgb]{ .867,  .922,  .969}  Sbd & \cellcolor[rgb]{ .867,  .922,  .969}  Rwy & \cellcolor[rgb]{ .867,  .922,  .969}  Bot & \cellcolor[rgb]{ .867,  .922,  .969}  Tc & \cellcolor[rgb]{ .867,  .922,  .969} mIoU \\
SAN (CVPR'23)~\cite{xu2023san} & 69.3  & 44.7  & 67.6  & 87.1  & 101.5  & 61.2  & 84.6  & 21.2  & 18.2  & 10.8  & 4.7   & 6.0   & 13.2  & 0.4   & 36.4  & 52.5  & 15.5  & 51.0  & 43.6  & 29.4  & 27.1  & 20.2  & 31.0  & 42.0  & 18.0  & 38.6  & 1.0   & 16.0  & 10.5  & 39.8  & 35.4  \\
\bottomrule
\end{tabular}%
}
 	\setlength{\abovecaptionskip}{0.00 cm}
  \caption{Per-class results for OVDG-SS in CS-7→Mapi-30.}
  \label{Tab6_ovdg_per_class_30}%
  \vspace{-0.2cm}
\end{table*}%

\begin{table*}[!t]
  \centering
   \renewcommand{\arraystretch}{1.05}
  \resizebox{0.795\textwidth}{!}{
\begin{tabular}{cccccccccccc}
\toprule
\multicolumn{12}{c}{\textbf{CS-7→RW-10 ViT-B/16}} \\
\midrule
      & \cellcolor[rgb]{ .886,  .937,  .855}\textbf{Rd} & \cellcolor[rgb]{ .886,  .937,  .855}\textbf{ Sw} & \cellcolor[rgb]{ .867,  .922,  .969} \textbf{ Bar} & \cellcolor[rgb]{ .867,  .922,  .969} \textbf{ PV} & \cellcolor[rgb]{ .867,  .922,  .969} \textbf{ WV} & \cellcolor[rgb]{ .867,  .922,  .969} \textbf{ PO} & \cellcolor[rgb]{ .867,  .922,  .969} \textbf{ Wkr} & \cellcolor[rgb]{ .867,  .922,  .969} \textbf{ Con} & \cellcolor[rgb]{ .867,  .922,  .969} \textbf{ ABd} & \cellcolor[rgb]{ .867,  .922,  .969} \textbf{ TTC} & \cellcolor[rgb]{ .867,  .922,  .969} \textbf{mIoU} \\
\midrule
SAN (CVPR'23)~\cite{xu2023san} & 80.3  & 37.9  & 58.7  & 3.8   & 41.6  & 28.7  & 32.4  & 6.8   & 3.7   & 12.7  & 30.7  \\
CAT-Seg (CVPR'24)~\cite{cho2024cat} & 89.9  & 62.8  & 57.3  & 20.9  & 48.1  & 22.9  & 24.7  & 10.0  & 4.5   & 24.2  & 36.5  \\
CLIPSelf (ICLR'24)~\cite{wu2023clipself} & 88.1  & 62.9  & 58.8  & 20.0  & 49.5  & 23.5  & 24.8  & 9.5   & 4.5   & 23.8  & 36.5  \\
RSC-CLIPSelf (ICLR'25)~\cite{qiu2025refining} & 85.6  & 61.8  & 54.9  & 20.9  & 46.7  & 23.5  & 23.4  & 9.8   & 4.3   & 23.4  & 35.4  \\
MAFT+(ECCV'24)~\cite{jiao2024collaborative} & 89.4  & 55.9  & 51.4  & 20.9  & 44.8  & 20.8  & 23.8  & 9.7   & 4.3   & 23.8  & 34.5  \\
ESC-Net (CVPR'25)~\cite{ESCNet_2025_CVPR} & 91.7  & 59.9  & 54.9  & 21.0  & 48.2  & 21.8  & 24.7  & 9.8   & 4.3   & 24.8  & 36.1  \\
\bottomrule
\end{tabular}%
}
 	\setlength{\abovecaptionskip}{0.00 cm}
  \caption{Per-class results for OVDG-SS in CS-7→RW10.}
  \label{Tab7_ovdg_per_class_10}%
  \vspace{-0.2cm}
\end{table*}%

\newpage

{\small
\bibliographystyle{ieee_fullname}
\bibliography{ref}
}

\end{document}